\documentclass{article}

\usepackage[preprint]{acl}
\usepackage{makecell}
\usepackage{amsmath}
\usepackage{amssymb}
\usepackage{graphicx}
\usepackage{booktabs}
\usepackage{xcolor}
\usepackage{hyperref}
\usepackage{placeins}
\usepackage{microtype}
\usepackage{multirow}
\usepackage{tikz}
\usepackage{pdflscape}
\usetikzlibrary{decorations.pathreplacing,decorations.pathmorphing}
\usepackage{comment}
\usepackage{tabularx}

\begin{document}

\title{Prompt Compression via Activation Aggregation}
 
\author{
    Thibaud Ardoin, \hspace{0.5em} Semira Einsele, \hspace{0.5em} Evis Bregu, \hspace{0.5em} Gerhard Wunder \\
    Freie Universität Berlin \\
    \texttt{thibaud.ardoin@fu-berlin.de}
}
 
\maketitle
 
\begin{abstract}
Large language models process prompts by propagating activations through dozens of layers before generating a response. 
We ask whether the task-relevant information contained in an instruction prompt can be compressed into a single activation vector and re-injected into the model, replacing the original token sequence? 
We show this is achievable using a learned weighted sum of activations extracted at an intermediate layer and injected at an early layer of the target LLM. 
The compressed vector preserves task-relevant information, incurring an accuracy drop of under $2\%$ relative to full prompt processing. 
Beyond its practical implications, including reducing per-query computation for fixed instruction prompts without reprocessing the original token sequence, our analysis reveals structure in the activation space of LLMs: 
(i) mid-layer representations transfer meaningfully to early layers, suggesting a degree of cross-layer compatibility in how information is encoded; 
(ii) a single activation vector encodes a quantifiable and recoverable amount of semantic information; 
(iii) a weighted sum of activations is a robust representation compressor. 
 \end{abstract}

\section{Introduction} \label{sec:intro}

Modern LLMs are often queried with repeated prompt prefixes, such as instructions, system prompts, or few-shot examples. 
In naive inference, these tokens are recomputed at every call, even when the prefix remains fixed and only the user query changes. 
This is computationally wasteful: the same prefix is repeatedly tokenized, embedded, and propagated through the transformer layers. 
A natural question is whether the task-relevant information contained in such a prefix can be pre-computed, compressed, and reused directly in activation space.

Current systems already exploit repeated prompts through mechanisms such as KV-caching~\cite{pope2023efficiently,kwon2023efficient}, which store intermediate states associated with a fixed prefix.
These methods provide exact or near-exact reuse: the prefix computation is retained so that it does not need to be recomputed. 
In contrast, we ask a more restrictive question: can the effect of a prompt be compressed into a single prompt-dependent activation vector and then injected back into the model in a way that preserves its behavior?

\begin{figure}[t!]
  \centering
\hspace*{-0.2\linewidth}
  \resizebox{1.2\linewidth}{!}{\usetikzlibrary{arrows.meta, positioning, fit, backgrounds, decorations.pathreplacing}
\usetikzlibrary{positioning}

\definecolor{clrBlue}{HTML}{2271B3}
\definecolor{clrOrange}{HTML}{E87722}
\definecolor{clrPink}{HTML}{C03060}
\definecolor{clrTok}{HTML}{EFEFEF}
\definecolor{clrDark}{HTML}{2A2A2A}
\definecolor{clrMid}{HTML}{888888}

\definecolor{paletteDeepBlue}{HTML}{011959}
\definecolor{paletteOrange}{HTML}{F29D6D}
\definecolor{palettePink}{HTML}{FACCFA}
\definecolor{paletteGreen}{HTML}{5A7745}
\definecolor{paletteGreen2}{HTML}{226061}
\definecolor{paletteOrange2}{HTML}{DD954D}

\begin{tikzpicture}[
  font=\small\sffamily,
  >=Stealth,
  tok/.style={
    draw=palettePink!45, fill=palettePink,
    rounded corners=3pt,
    minimum height=0.62cm,
    inner xsep=8pt, inner ysep=4pt,
    font=\small\sffamily,
    align=center,
  },
  llmbox/.style={
    draw=clrBlue, fill=clrBlue!7,
    rounded corners=6pt,
    minimum height=3cm, minimum width=5cm,
    line width=1.3pt,
    align=center,
  },
  compbox/.style={
    draw=paletteGreen2, fill=paletteGreen2!20,
    rounded corners=5pt,
    minimum height=1.5cm, minimum width=2.6cm,
    line width=1.0pt,
    font=\small\sffamily,
    align=center,
  },
  patchbox/.style={
    draw=paletteOrange2, fill=paletteOrange2!30,
    rounded corners=4pt,
    minimum height=0.68cm, minimum width=1.3cm,
    line width=1.1pt,
    font=\small\bfseries,
    text=clrPink!85!black,
    align=center,
  },
  act/.style={
    draw=paletteOrange2!65, fill=paletteOrange2!18,
    minimum height=0.30cm, minimum width=0.92cm,
    rounded corners=1.5pt,
    line width=0.6pt,
  },
  arr/.style      ={->, line width=0.75pt, color=clrDark!65},
  arrpink/.style  ={->, line width=1.3pt,  color=paletteOrange2},
  arrorange/.style={->, line width=0.9pt,  color=clrOrange!80!clrDark},
  stepfont/.style ={font=\large\bfseries\sffamily, text=clrDark},
  sublabel/.style ={font=\tiny\sffamily, color=clrMid},
]



\node[tok] (i1) at (-0.9, -1) {$t_1$};
\node[tok, right=4pt of i1] (i3) {\small$\cdots$};
\node[tok, right=4pt of i3] (i4) {$t_n$};
\node[sublabel, above=20pt of i3.north, anchor=north] {\large information prompt};

\node[stepfont] (hdr1) at ([yshift=1.3cm]i3.north) {\underline{\textbf{Step 1}: Extraction}};

\node[llmbox, below=0.7cm of i3] (llm1) {};
\node[font=\small\bfseries, text=clrBlue] at ([xshift=-1.9cm, yshift=+0.38cm]llm1.south) {LLM};
\node[font=\tiny\sffamily, text=clrBlue!65] at ([xshift=-1.9cm, yshift=+0.8cm]llm1.south) {(frozen)};
\foreach \y in {0.60, 0.0, -0.60}{
  \draw[clrBlue!22, line width=0.45pt]
    ([yshift=\y cm]llm1.west) -- ([yshift=\y cm]llm1.east);
}

\node[act, above=-40pt of llm1] (a2) {};
\node[act, left=4pt of a2, xshift=0pt] (a1) {};
\node[act, right=4pt of a2, xshift=0pt] (a3) {};
\node[sublabel, below=20pt of a2, anchor=south] {\large hidden states};

\draw[arr] (i1.south) -- (a1.north);
\draw[arr] (i3.south) -- (a2.north);
\draw[arr] (i4.south) -- (a3.north);




\coordinate[right=100pt of a3] (ang)  ;

\node[act, below=25pt of ang] (aa2) {};
\node[act, left=4pt of aa2, xshift=0pt] (aa1) {};
\node[act, right=4pt of aa2, xshift=0pt] (aa3) {};

\node[stepfont, above=45pt of aa2] (hdr2) {\underline{\textbf{Step 2}: Compression}};

\node[compbox, below=30pt of aa2] (comp) {Hidden State \\ Compressor};

\node[patchbox, below=1.3cm of comp] (patch) {\textbf{patch}};

\draw[arrorange, line width=1.4pt, rounded corners=10pt] (a3.east) -| (aa2.north) node[font=\large\sffamily, pos=0.3, above, color=clrMid] {extraction};;

\draw[arrorange] (aa2.south) -- (comp.north);
\draw[arrorange] (aa1.south) -- ([xshift=-0.5cm]comp.north);
\draw[arrorange] (aa3.south) -- ([xshift=0.5cm]comp.north);

\draw[arrpink] (comp.south) -- (patch.north);

\node[stepfont] (hdr3) at (0, -6.4) {\underline{\textbf{Step 3}: Injection}};

\node[tok] (q3) at (0, -8) {Placeholder};
\node[tok, left=4pt of q3] (q1)  {$<$BoS$>$};
\node[tok, right=4pt of q3] (q2) {Question};
\node[sublabel, above=20pt of q3.north, anchor=north] {\large query prompt};

\node[llmbox, below=0.7cm of q3] (llm2) {};
\node[font=\small\bfseries, text=clrBlue] at ([xshift=-1.9cm, yshift=+0.38cm]llm2.south) {LLM};
\node[font=\tiny\sffamily, text=clrBlue!65] at ([xshift=-1.9cm, yshift=+0.8cm]llm2.south) {(frozen)};
\foreach \y in {0.60, 0.0, -0.60}{
  \draw[clrBlue!22, line width=0.45pt]
    ([yshift=\y cm]llm2.west) -- ([yshift=\y cm]llm2.east);
}

\node[patchbox, below=40pt of q3] (patchSlot) {\textbf{patch}};
\node[act, right=4pt of patchSlot] (actSlot1) {};
\node[act, left=4pt of patchSlot] (actSlot2) {};

\draw[arr] (q1.south)  -- ++(0,-0.2) -|([xshift=0cm]actSlot2.north);
\draw[arr] (q2.south)  -- ++(0,-0.2) -| ([xshift=0cm]actSlot1.north);
\draw[arr] (q3.south) -- (patchSlot.north);

\node[tok, below=2.5cm of actSlot1.south] (ans) {Answer};
\node[sublabel, below=6pt of ans, anchor=north] {\large about information prompt};
\draw[arr] (actSlot1.south) -- (ans.north);

\draw[arrorange, line width=1.4pt, rounded corners=10pt]
  (patch.south)
    -- (patch |- patchSlot)
    -- (patchSlot.east) node[font=\large\sffamily, pos=0.3, above, color=clrMid] {injection};

\end{tikzpicture}}
  \caption{
  Overview of the proposed three-step framework: extract hidden states from an input prompt, compress them into a patch vector, and inject the patch vector into  a placeholder token to answer a query without direct access to the original prompt.
}
  \label{fig:intro-representation}
\end{figure}
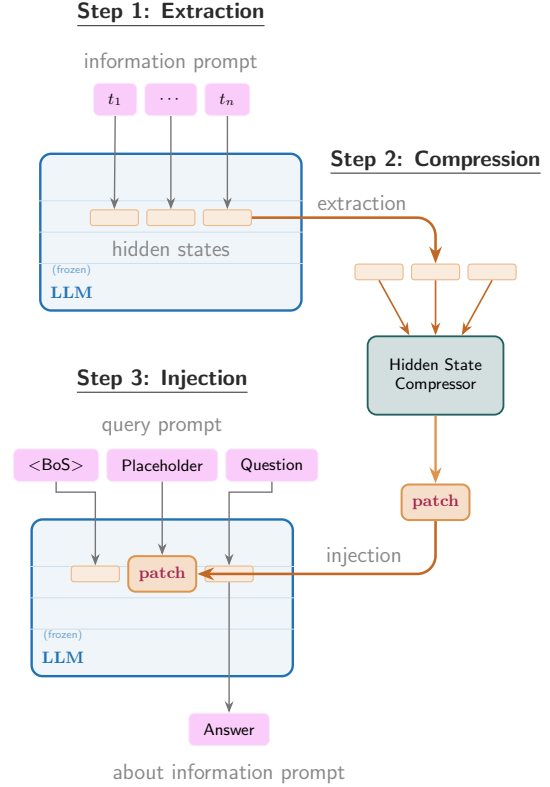

Beyond the engineering motivation, this question probes a more fundamental issue: \emph{how is task-relevant information from a prompt represented in the activation space of an LLM?} 
Prior work on activation engineering has shown that steering vectors can influence complex model behavior through simple additions in latent space~\cite{subramani2022extracting}. 
More generally, the linearity hypothesis suggests that some high-level concept representations allow simple arithmetic operations~\cite{Mikolov2013linguistic,liu2023incontext}.

With respect to information compression, task vectors~\cite{hendel2023incontext} condense in-context learning examples, but they do not extend naturally to general information prompts.
Recent methods have shown that long contexts or arbitrary prompts can be compressed into a small number of tokens or representations~\cite{mu2023learning,ge2023context}. 
However, these strong results typically rely on fine-tuning the full LLM or on extensive training procedures before the compression behavior emerges.

This leaves open the question of whether context-relevant prompts can be compressed for off-the-shelf LLMs using a lightweight method that requires only minimal training.

To address this gap, we propose a simple activation-space compression method based on a weighted sum. 
This design is motivated by mechanistic interpretability: if task-relevant information is represented in internal activations, then it may be possible to compress prompts directly in latent space with minimal overhead.
As illustrated in~\autoref{fig:intro-representation}, the method takes as input a prompt $p=[t_1,\ldots,t_T]$ that encodes relevant information or a task description. 
We extract the hidden-state sequence $H^{(m)}(p)\in\mathbb{R}^{T\times d}$ at an intermediate layer $m$ and compress it into a prompt-dependent patch vector $v\in\mathbb{R}^d$. 
In a second forward pass, the model receives a shortened input consisting of a \textit{placeholder token}, whose activation is overwritten by $v$, followed by a query about the original prompt $p$. The model must then generate a response without direct access to $p$.
We find that the best compression function is a weighted sum over the hidden states in $H^{(m)}(p)$, with weights predicted by a small learned multilayer perceptron (MLP). This inherently lossy compression reduces test accuracy by only 2\% on task-instruction prompts. 
Surprisingly, the simple weighted-sum compressor consistently outperforms the heavier end-to-end trained Transformer compressor in our main experiments.

In this paper, we make the following contributions:
\begin{enumerate}
    \item We propose a two-pass framework for activation-space prompt compression, in which a prompt is compressed into a single patch vector and later re-injected into the model through a placeholder token.

    \item We show that a lightweight MLP can learn a weighting function that constructs patch vectors which generalize to unseen prompts and tasks.
    
    \item Through a layer-wise analysis, we uncover a new pattern in activation engineering: best performances are obtained when information is extracted from a middle layer and injected into an early layer.

    \item We provide a detailed analysis of the resulting patch vectors, along with ablations over key design choices.

    \item We release a \textit{Toy Task} dataset and the code required to reproduce our results at \href{https://anonymous.4open.science/r/compressed-semantic-212F/README.md}{our anonymous repository}.
\end{enumerate}

\section{Method} \label{sec:method}

Given a prompt $p = [t_1, \ldots, t_T]$ and an LLM with $L$ attention layers, a hidden-state sequence 
$$H^{(m)}(p)=\bigl(h^{(m)}_1, \ldots, h^{(m)}_T \bigr) \in \mathbb{R}^{T \times d}$$
can be extracted at an intermediate extraction layer $m \in \{1, \dots, L\}$. 
Here $h_i^{(m)} \in \mathbb{R}^d$ denotes the hidden-state vector downstream of the token $t_i$ at position $i$ at layer $m$.

A \textit{compression function} $f: \mathbb{R}^{T\times d} \rightarrow \mathbb{R}^{d}$ maps the sequence of hidden-states to a single \textit{patch vector}. 
$$v = f\bigl(H^{(m)}(p)\bigr) \in \mathbb{R}^d.$$

As depicted in~\autoref{fig:intro-representation}, this vector is then injected at an early injection layer $e < m$, at position $k\in\{1, \dots, T\}$, replacing the hidden-state downstream of the \textit{placeholder} token $t_k$. The model then continues its forward pass from layer $e$ onward and generates a response. 
This design is motivated by prior work suggesting that intermediate layers contain richer semantic representations~\cite{panickssery2023steering}, as well as by our empirical observation that early injection performs better (see~\autoref{sec:exp:ablation} and~\autoref{fig:layer-heatmap}). 
One possible explanation is that injecting the compressed representation earlier leaves more transformer layers for it to be processed and integrated before prediction.

\begin{figure}[t]
\centering
\resizebox{\columnwidth}{!}{%
\begin{tikzpicture}

\definecolor{paletteDeepBlue}{HTML}{011959}
\definecolor{paletteOrange}{HTML}{F29D6D}
\definecolor{palettePink}{HTML}{FACCFA}
\definecolor{paletteGreen}{HTML}{5A7745}
\definecolor{paletteGreen2}{HTML}{226061}
\definecolor{paletteOrange2}{HTML}{DD954D}

\tikzstyle{act}=[
    draw=paletteOrange,
    fill=paletteOrange!12,
    rounded corners=2pt,
    minimum width=1cm,
    minimum height=0.6cm,
    align=center,
    font=\large,
]

\tikzstyle{patch}=[
    draw=paletteOrange2,
    fill=paletteOrange2!30,
    line width=1.1pt,
    rounded corners=2pt,
    minimum width=0.9cm,
    minimum height=0.45cm,
    align=center,
    font=\large,
]

\tikzstyle{module}=[
    draw=paletteGreen2, 
    fill=paletteGreen2!20,
    rounded corners=3pt,
    minimum width=1cm,
    minimum height=0.8cm,
    align=center,
    font=\large,
]

\tikzstyle{bigmodule}=[
    draw=paletteGreen2, 
    fill=paletteGreen2!20,
    rounded corners=3pt,
    minimum width=4.2cm,
    minimum height=0.8cm,
    align=center,
    font=\large,
]

\tikzstyle{weight}=[
    draw=palettePink,
    fill=palettePink!40,
    rounded corners=2pt,
    minimum width=0.65cm,
    minimum height=0.4cm,
    align=center,
    font=\large,
]

\tikzstyle{sum}=[
    draw,
    circle,
    minimum size=0.55cm,
    align=center,
    font=\large,
    fill=gray!10
]


\node[font=\normalsize\bfseries] (titleA) at (3.0,4.0) {(a) Weighting MLP};

\node[act] (hdots) at ([xshift=-2cm, yshift=-0.7cm]titleA.south) {$\cdots$};
\node[act] (h2) at ([xshift=-1.2cm]hdots) {$h_2^{(m)}$};
\node[act] (h1) at ([xshift=-1.2cm]h2) {$h_1^{(m)}$};
\node[act] (hT) at ([xshift=1.2cm]hdots) {$h_T^{(m)}$};

\node[module] (m1) at ([yshift=-1cm]h1.south) {$M_\theta$};
\node[module] (m2) at ([yshift=-1cm]h2.south) {$M_\theta$};
\node[module] (mT) at ([yshift=-1cm]hT.south) {$M_\theta$};

\node[weight] (w1) at ([yshift=-1cm]m1.south) {$w_1$};
\node[weight] (w2) at ([yshift=-1cm]m2.south) {$w_2$};
\node[weight] (wdots) at ([xshift=1.2cm]w2) {$\cdots$};
\node[weight] (wT) at ([yshift=-1cm]mT.south) {$w_T$};

\draw[->, thick] (h1) -- (m1);
\draw[->, thick] (h2) -- (m2);
\draw[->, thick] (hT) -- (mT);

\draw[->, thick] (m1) -- (w1);
\draw[->, thick] (m2) -- (w2);
\draw[->, thick] (mT) -- (wT);

\node[sum] (sumA) at ([xshift=1cm]mT.east) {$\sum$};
\node[patch, anchor=west] (vA) at ([xshift=0.5cm]sumA.east) {$v^{(m)}=\sum_{i=1}^{T} w_i h_i^{(m)}$};

\draw[->] (hT.east) -- (sumA.north);
\draw[->] (wT.east) -- (sumA.south);
\draw[->, thick] (sumA.east) -- (vA.west);

\node[font=\large,  align=center] (legend1) at ([yshift=-0.5cm, xshift=5cm]w1.south) {
Each $w_i \in \mathbb{R}$ is computed from a single activation $h_i^{(m)}\in\mathbb{R}^d$  
};


\begin{scope}[yshift=-4.6cm]

\node[font=\normalsize\bfseries, align=left] (titleB) at ([yshift=-0.5cm]legend1.south) {(b) Transformer Compressor};


\node[act] (thdots) at ([yshift=-0.7cm]titleB.south) {$\cdots$};
\node[act] (th2) at ([xshift=-1.2cm]thdots) {$h_2^{(m)}$};
\node[act] (th1) at ([xshift=-1.2cm]th2) {$h_1^{(m)}$};
\node[act] (thT) at ([xshift=1.2cm]thdots) {$h_T^{(m)}$};

\node[bigmodule] (trans) at ([yshift=-1cm, xshift=-0.5cm]thdots.south) {$E_\phi$ over full sequence};

\node[patch] (patch) at ([yshift=-1cm]trans.south) {Compressed \\ Patch};


\draw[->, thick] (th1) -- (trans.north -| th1);
\draw[->, thick] (th2) -- (trans.north -| th2);
\draw[->, thick] (thdots) -- (trans.north -| thdots);
\draw[->, thick] (thT) -- (trans.north -| thT);




\draw[->, thick] (trans.south) -- (patch.north);


\node[font=\large, align=center] at ([yshift=-0.7cm]patch.south) {
Direct learning of the activation compression:\\ $H^{(m)}(p) \in \mathbb{R}^{d\times n} \rightarrow v \in \mathbb{R}^d$.
};

\end{scope}


\end{tikzpicture}%
}

\caption{
    Comparison of our two activation-compression methods: the Weighting MLP and the Transformer Compressor. The W-MLP scores each contextualized activation separately, while the TC consumes the full sequence $H^{(m)}(p)$ to produce a patch vector in $\mathbb{R}^d$.
}
\label{fig:mlp-vs-transformer}
\end{figure}
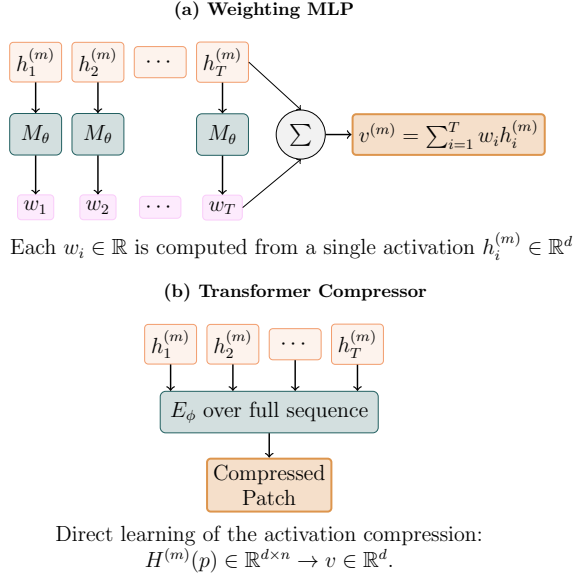

We propose two compression functions $f$, depicted in~\autoref{fig:mlp-vs-transformer}.
The first, the \textit{Weighting MLP} (W-MLP), operates under the strong assumption that activations can be compressed via a simple weighted sum. 
The second, the \textit{Transformer Compressor} (TC), directly optimizes the compression of activations into the patch vector $v$ with end-to-end training. 
Both methods are trained with cross-entropy between the target output of the LLM and its logits after being patched by $v=f(H^{(m)}(p))$. The hyperparameters of all training procedures have been optimized with grid search, details are given in~\autoref{app:hyperparameters}.

\subsection{Weighting MLP} \label{sec:method:mlp}

The linear representation hypothesis suggests that the internal representations of two concepts can be summed to superimpose their effects~\cite{Mikolov2013linguistic}. 
Following these considerations, we hypothesize that the information contained in multiple activation vectors can be compressed through a weighted sum. These weights can be hand-designed, as explored in a case study presented in~\autoref{app:hand-made-weights}. However, such hand-crafted weights are imprecise and fail to generalize beyond a specific prompt. To eliminate this manual choice, we train an MLP to predict the weights from the activations themselves. 
Concretely, we use a network
$$
M_\theta: \mathbb{R}^d \to \mathbb{R}
$$
that maps each intermediate activation $h_i^{(m)}$ to a scalar weight $w_i$ used to form the patch vector:
\begin{equation}
 M_\theta\!\bigl(h^{(m)}_i\bigr) = w_i, \qquad  v = \sum_{i=1}^{T} w_i\, h^{(m)}_i.
    \label{eq:mlp-weighting}
\end{equation}

The MLP does not directly generate the full patch vector. Instead, it learns how strongly each token activation should contribute to the final compressed representation.

In our implementation, $M_\theta$ is a small feed-forward network with 4 layers with hidden dimension $[2048, 1024, 512, 256]$, and a ReLU activation function.

\subsection{Transformer Compressor} \label{sec:method:transformer}

To explore the limits of compression, we train a transformer encoder $E_\phi$ that attends over the full sequence of hidden representations and produces a single compressed vector $v$ via a \texttt{[CLS]}-style token:

\begin{equation}
\begin{aligned}
E_\phi : \mathbb{R}^{T \times d} &\to \mathbb{R}^{d}, \\
H^{(m)} = [h^{(m)}_1,\ldots,h^{(m)}_T]
&\mapsto v = E_\phi(H^{(m)})_{\texttt{CLS}} .
\end{aligned}
\end{equation}

The hidden dimension of $E_\phi$ is $64$, with $2$ attention heads and $2$ layers. 
The Transformer Compressor is an end-to-end solution that does not impose any prior structure on the patch representation. 
While this makes it more expressive than the W-MLP, we show that this makes it prone to overfitting on specific tasks, as discussed in~\autoref{sec:exp:accuracy}.

\section{Experiments} \label{sec:setup}

We evaluate whether relevant information contained in a prompt can be compressed into a single activation-space patch vector. Our experiments are designed to answer four research questions: 
(1) Does the compressed vector preserve the behavior induced by the original prompt? 
(2) How does performance change as the complexity of the compressed information increases? 
(3) How do the W-MLP and TC compare? and 
(4) How well do these models generalize beyond their training distribution?

\subsection{Experimental Setup}

\paragraph{Open-weight LLMs.}
We conduct our experiments with \texttt{Llama3.1-8B-Instruct}~\cite{Grattafiori2024}, chosen for its broad availability and favorable performance-to-size trade-off. To evaluate the generality of our approach across model families and scales, we further test \texttt{Ministral3-8B-Instruct-2512}~\cite{liu2026ministral}, a strong 8B-parameter model, \texttt{Qwen3.5-4B}~\cite{qwen3.5}, and \texttt{Llama3.2-1B-Instruct}. All experiments are run on an \texttt{NVIDIA RTX A5500}.

\definecolor{compressblue}{RGB}{173, 216, 230}   
\definecolor{injectorange}{RGB}{255, 213, 170}    

\definecolor{paletteDeepBlue}{HTML}{011959}
\definecolor{paletteOrange}{HTML}{F29D6D}
\definecolor{palettePink}{HTML}{FACCFA}
\definecolor{paletteGreen}{HTML}{5A7745}
\definecolor{paletteGreen2}{HTML}{226061}
\definecolor{paletteOrange2}{HTML}{DD954D}

\begin{table}[t]
\centering
\setlength{\tabcolsep}{4pt}

\begin{tabular}{p{\linewidth}}
\toprule
\textbf{Toy Task}  \\
\midrule

\textit{Extraction:} \\ 
\colorbox{palettePink}{\small \texttt{Given a country, return its capital.}} \\
\vspace{0.1em}
\textit{\normalsize Injection:} \\
\small 
\colorbox{paletteGreen2!50}{\texttt{¿}}{\texttt{Italy:}} \colorbox{paletteOrange!60}{\texttt{Rome}}\\

\midrule
\textbf{ARC-Easy}  \\
\midrule
\textit{Extraction:} \\
\colorbox{palettePink}{\small \texttt{Which of the following is a primary color?}}\\ 
\vspace{0.1em}
\textit{\normalsize Injection:} \\
\small \texttt{Reply with ONLY the letter of the correct answer (A, B, C, or D)}
\colorbox{paletteGreen2!50}{\texttt{¿}} \texttt{(A) Green (B) Orange (C) Red (D) Purple} \\
\texttt{Answer: \colorbox{paletteOrange!60}{C}} \\
\bottomrule
\end{tabular}

\caption{Extraction and injection prompts for each dataset. \colorbox{palettePink}{Pink} indicates the part encoded into the compressed representation $v$. \colorbox{paletteGreen2!50}{Green} indicates the placeholder token where $v$ is patched into the residual stream. \colorbox{paletteOrange!60}{Orange} indicates the generated text by the LLM. In this representation the special tokens are excluded for clarity.}

\label{tab:prompts}
\end{table}

\paragraph{Data.}

We evaluate our compression functions on two types of 
data: a set of homemade custom tasks and a standardized 
benchmark.

The \textbf{Toy Task} dataset is inspired by the task-vector experiments of~\citet{hendel2023incontext}. However, instead of representing tasks through examples, we specify them directly through instructions. 
We construct 11 knowledge-retrieval tasks, each defined by a simple input-output mapping (e.g., given a \texttt{country}, predict its \texttt{capital}). 
For each task, we create 10 prompt templates for training and 10 for evaluation, with approximately 100 (\texttt{entity}, \texttt{target}) pairs for training and 20 for evaluation per task.

This dataset provides a controlled setting in which each task is precisely defined and correctness can be verified by exact string matching. It also makes it straightforward to train the compression functions using a cross-entropy loss over the LLM output logits. Details and examples of the dataset are provided in~\autoref{app:datasets}.

The second dataset is \textbf{ARC-Easy}~\cite{clark2018think}, a multiple-choice benchmark in which each question has four answer candidates (\texttt{A}, \texttt{B}, \texttt{C}, \texttt{D}). 
We compress the question itself into the representation $v$, while the model receives only the answer choices, as depicted in~\autoref{tab:prompts}. 
A failed compression leaves the model without access to the question, resulting in near-chance performance, whereas successful compression should recover accuracy close to the uncompressed baseline. 
As in the Toy Task dataset, the multiple-choice format allows direct training with cross-entropy loss over the four candidate logits. 
We choose ARC-Easy because its questions are relatively short and the required knowledge is largely covered by the capabilities of modern LLMs. 
This allows us to evaluate the compression strategy without prompt length or question complexity becoming the main limiting factors, unlike in more verbose and technical benchmarks such as MMLU~\cite{hendrycks2020measuring}.

\paragraph{Extraction and Injection Layers.}
Unless otherwise stated, we use layer $m=12$ for extraction and layer $e=2$ for injection in \texttt{Llama3.1-8B-Instruct}. These settings were chosen based on a preliminary ablation over extraction-injection layer pairs and are treated as model-dependent hyperparameters. 
The full ablation study is reported in~\autoref{sec:exp:ablation}.

\paragraph{Injection Token.}
The patch vector $v$ is injected by replacing the activation downstream of a designated \textit{placeholder token} at layer $e$. We use the UTF-8 replacement character \texttt{U+FFFD} (rendered as~\texttt{¿} here) as the placeholder, because it is unlikely to evoke semantic associations in the model's vocabulary and is available across all models. 
It is a single character that is consistently mapped to a single token by the tokenizer, regardless of context. 
We study the effect of this token choice and the injection position in~\autoref{sec:exp:ablation}.

\paragraph{Baselines.}
We compare our compression methods against two baselines: 
(i) the \textit{full-prompt baseline}, which gives the model access to the original task or question without compression and serves as an upper bound, and
(ii) the \textit{masked-prompt baseline}, which preserves the prompt structure but replaces the compressed content with the placeholder token and does not apply any patching, serving as a lower bound. Together, these baselines bracket the performance of our compressed representations.

\subsection{Compression of Tasks Based on Instruction Prompts.} \label{sec:exp:accuracy}

\begin{table}[h]
    \centering
    \small


        \begin{tabular}{lrrrr}
    \toprule
        Method & Train & Test & \makecell{OOD \\ Tasks} \\
        \midrule
        Masked Prompt & $32.55$ & $34.03$ & $18.45$ \\
        TC            & $88.87$ & $70.63$ & $43.89$ \\
        W-MLP         & $89.23$ & $85.35$ & $63.01$ \\
        Full prompt   & $85.75$ & $86.92$ & $94.95$ \\
        \bottomrule
    \end{tabular}
    \caption{Exact-match accuracy on the training and test sets of the Toy Task Dataset for the Weighting MLP (W-MLP), the Transformer Compressor (TC), and both baselines, using \texttt{Llama-3.1-8B-Instruct}.}
\label{tab:main-toy-results}
\end{table}

We first evaluate whether a single patch vector can preserve the behavior induced by the original prompt. 
We train the W-MLP and TC on eight task families from the Toy Task dataset: capitals, ISO country codes, continents, event years, English-to-French translation, art authorship, antonyms, and currency codes. We reserve three tasks for out-of-distribution evaluation: French-to-English translation, chemical element, and tuple addition. For each task family, we compare our compression methods against the two baselines defined above.

The resulting accuracies for \texttt{Llama-3.1-8B-Instruct} are reported in~\autoref{tab:main-toy-results}. 
The gap between the masked- and full-prompt baseline accuracies confirms that the task prompt contains important information, and that the entity alone is not sufficient for most tasks. 
We observe that both methods consistently outperform the masked-prompt baseline and recover a substantial portion of the gap to the full-prompt baseline, especially on in-distribution tasks.

\begin{figure}[t]
    \centering
    \includegraphics[width=\linewidth]{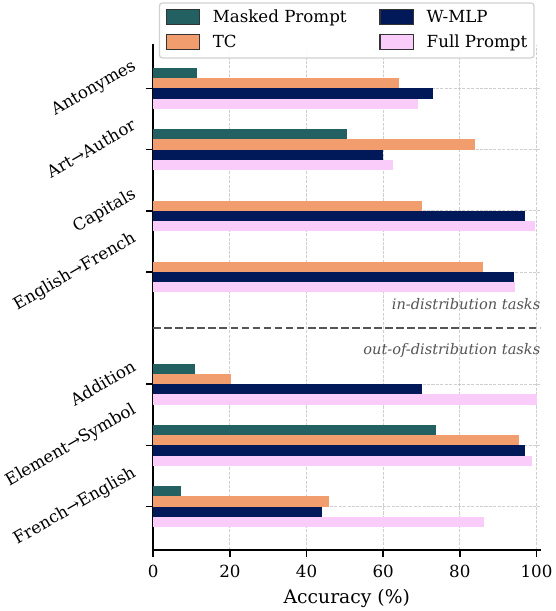}
    \caption{Comparison of test performance for a sub-sample of the in-distribution Tasks and out-of-distribution Tasks.}
    \label{fig:accuracy-task-comparaison}
\end{figure}

We observe that the simple W-MLP method outperforms the end-to-end TC solution, suggesting that much of the task-relevant information can be manipulated through simple linear operations in representation space. 
On in-distribution tasks, W-MLP performs within $2\%$ of the full-prompt baseline, suggesting that a single compressed vector can preserve most of the prompt instruction. 
As shown in~\autoref{fig:accuracy-task-comparaison}, its OOD performance varies substantially by task and degrades more noticeably on harder tasks.

In contrast, the TC performs substantially worse on the test set, reaching only $70.63\%$ accuracy. 
Since this model is trained directly to reconstruct the patch vector, it may be more prone to overfitting. 
This hypothesis is supported by its high training accuracy and by the fact that it even outperforms the full-prompt baseline on the art-to-author task. 
Overall, these results suggest that the more expressive TC can learn task-specific bottleneck representations, but W-MLP generalizes better.

\subsection{Compression Beyond Task Prompts}

\begin{figure}[b]
    \centering
\includegraphics[width=\linewidth]{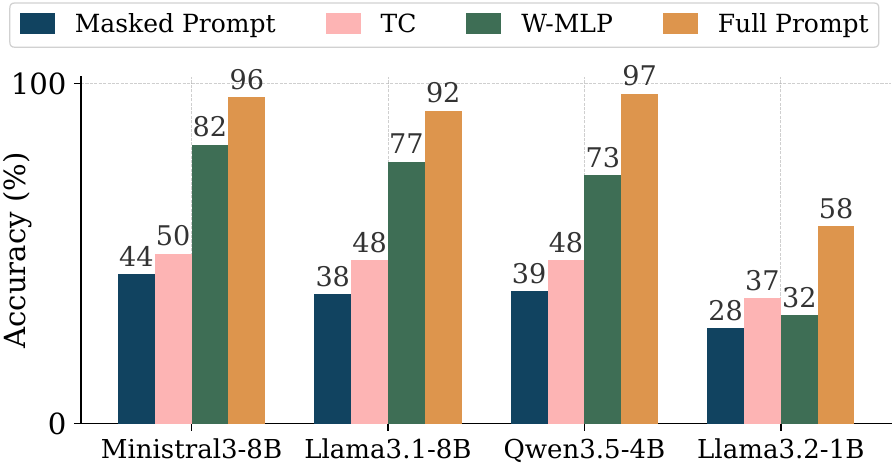}
    \caption{ARC-Easy test accuracy across LLMs, comparing the Transformer Compressor (TC), the Weighting MLP (W-MLP), and the full- and masked-prompt baselines}
    \label{fig:arceasy-accuracy-plot}
\end{figure}

ARC-Easy provides a stricter test of generalization than the out-of-distribution Toy Task setting, because each example contains a distinct one-shot prompt rather than a fixed task description. 
As a result, this benchmark evaluates whether the compression mechanism can generalize beyond a single prompt template while preserving information relevant for downstream multiple-choice prediction. 
Test accuracies across methods and LLMs are shown in~\autoref{fig:arceasy-accuracy-plot}.

Overall, W-MLP learns a more general compression function and achieves test accuracy close to the full-prompt reference upper bound on most models. 
In contrast, TC appears to overfit more strongly: although it can reach training accuracy comparable to the W-MLP, its test performance drops much more sharply. 
On \texttt{Llama3.1-8B}, W-MLP reaches $93\%$ training accuracy and $77\%$ test accuracy, whereas the TC reaches $87\%$ training accuracy but only $48\%$ test accuracy. 
This gap suggests again that W-MLP induces a simpler and more stable compression function, while TC has greater capacity but weaker generalization.

We also find that smaller models achieve lower test accuracy with W-MLP. 
\texttt{Llama3.2-1B} and \texttt{Qwen3.5-4B} have hidden dimensions of $1,024$ and $2,560$, respectively, compared to $4,096$ for both \texttt{Llama3.1-8B} and \texttt{Ministral3-8B}. 
This suggests that smaller hidden dimensions may limit the representational capacity available for compression, and that activation-space compression may improve as the dimensionality of the model's internal representations increases.

\subsection{Ablation: Extraction and Injection Layers} \label{sec:exp:ablation}

\begin{figure}[htbp]
    \centering
    \includegraphics[width=\linewidth]{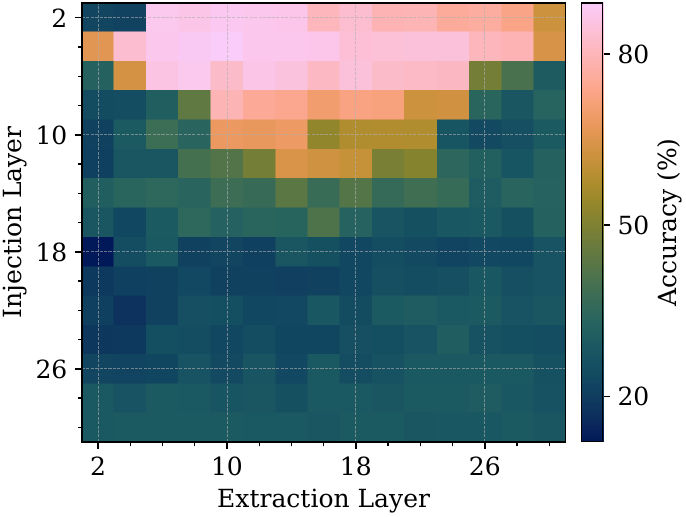}
    \caption{
    Heatmap of the Weighting MLP accuracy after 1 epoch on the Toy Tasks, according to extraction layer and injection layer.
    }
    \label{fig:layer-heatmap}
\end{figure}

We next ablate the extraction and injection layers to assess their influence on compression quality. \autoref{fig:layer-heatmap} reports the training accuracy of W-MLP after one epoch across different extraction-injection layer pairs on a subset of tasks.
This ablation reveals a clear trend: the best performance is obtained when activations are extracted from an intermediate layer and injected into an early layer.
The strong performance of intermediate-layer extraction is consistent with prior work showing that middle layers tend to encode more abstract and task-relevant information~\cite{Marks2023}.
The effectiveness of early injection may be explained by the fact that it allows more transformer blocks to process and integrate the compressed representation before prediction.
Combining a middle-layer extraction with an early-layer injection is also consistent with prior work suggesting that a model’s internal state remains interpretable across layers, rather than being localized to a single layer~\cite{elhoushi2024layerskip}.

The heatmap provides a coarse view of the layer-pair landscape and suggests a promising region around extraction layer 10 and injection layer 4. 
A finer search over this region identifies extraction at layer 12 and injection at layer 2 as the optimal configuration for \texttt{Llama3.1-8B}, which we use in all main experiments.

\subsection{Ablation: Placeholder Token and Patching Position}

\begin{figure}[t]
    \centering
    \includegraphics[width=\linewidth]{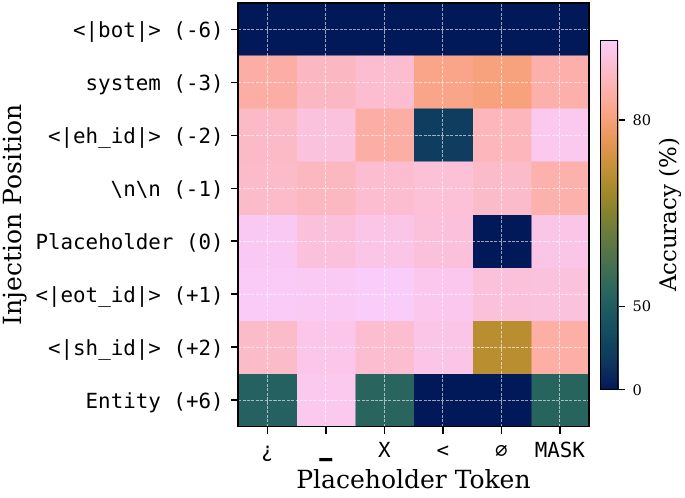}
    \caption{Heatmap of the W-MLP accuracy on the Toy Tasks as a function of the placeholder token and patching position. 
    The $y$-axis indicates the token position at which the patch vector is injected, relative to the placeholder position. The absence of a placeholder token is denoted by $\varnothing$.
    }
    \label{fig:placeholder-position}
\end{figure}

To the best of our knowledge, our specific activation-patching setting has not been previously explored, leaving several design choices open, including the choice of placeholder token and the exact position at which patching is applied. Results of the ablation study are shown in~\autoref{fig:placeholder-position}.
Our chosen configuration, patching downstream of token \texttt{¿}, performs competitively and ranks among the best settings, although the margin over nearby alternatives is small. This suggests limited sensitivity to the exact placeholder token, as long as it is a neutral and consistently tokenized choice.
In contrast, a clearer trend emerges along the position axis: patching further away from the placeholder token and toward either the beginning-of-text token (\texttt{<|bot|>}) or the assistant answer consistently degrades performance. This indicates that the intervention is most effective when applied close to the placeholder position.

These results also suggest room for improvement and motivate extensions that inject multiple patch vectors simultaneously, potentially providing richer information at the cost of a more careful positional design.

\section{What Do Patch Vectors Represent?} \label{sec:patching-vector-analysis}

The results in the previous section show that a learned patch vector can recover most of the performance of the original prompt. 
Here, we analyze what kind of information this vector captures. 
Specifically, we ask whether the patch vector remains close to the original prompt activations and whether the learned weighting mechanism focuses on semantically meaningful prompt tokens. 

To gain intuition on the compression mechanism, we qualitatively inspect the learned weights assigned to each token in the weighted sum in~\autoref{fig:weighting-prompt-qualitative}. 
A consistent pattern emerges: tokens carrying dense semantic content, such as keywords or named entities, receive systematically higher weights, while low-information tokens such as determiners are comparatively suppressed. 

To move beyond purely qualitative observations, we conduct a controlled experiment using 100 LLM-generated reformulations of the prompt \textit{``What is the industry this company operates in?''}. 
Each reformulation varies in surface form while preserving task semantics, allowing us to track which token consistently receives the highest learned weight across phrasings. In $85\%$ of cases, the highest weight is assigned to either \textit{industry}, \textit{field}, or \textit{sector}, directly reflecting the target of the retrieval task. This provides quantitative support for the qualitative pattern observed in~\autoref{fig:weighting-prompt-qualitative}: the compression mechanism identifies and prioritizes the semantically load-bearing token in the prompt. We note, however, that semantic salience is not formally defined here, so this analysis should be interpreted as a qualitative diagnostic rather than a formal measure of token importance.

In the remaining $10\%$ of cases, the highest weight is assigned to the terminal punctuation token. Strikingly, even when it does not rank first, the interrogation mark appears as the second highest weighted token in $75\%$ of cases across all reformulations. 
Rather than contradicting the semantic salience finding, this is consistent with the well-documented attention sink phenomenon in auto-regressive models, whereby punctuation and sentence-boundary tokens accumulate disproportionate attention weight as aggregators of contextual information~\cite{xiao2024efficient, chauhan2026punctuations}. 
The W-MLP thus appears to exploit the same representational structure that attention heads do, further grounding the compression mechanism in known properties of transformer activation space.
The focus on key semantic words and on punctuation is also consistent with the hand-crafted weights studied in~\autoref{app:hand-made-weights}.

These observations are further corroborated by two independent metrics: the KL-divergence between the compressed patch and individual token activations, and their cosine similarity. 
Both metrics assign highest proximity to the same semantically salient tokens identified by the learned weights. 
\autoref{fig:weighting-prompt-qualitative} confirms that this interpretability extends beyond the weights themselves to the final aggregated patch vector, which remains geometrically close to the activations of task-critical tokens. 
Additional examples are provided in~\autoref{app:weighting-examples}.

\definecolor{paletteBlue3}{HTML}{011959}
\definecolor{paletteOrange3}{HTML}{F29D6D}

\begin{table}[t]
    \centering
    \small
    \begin{tabularx}{\columnwidth}{X}

        \toprule
        \textit{(a) Weight attribution of W-MLP}\\
\setlength{\fboxsep}{1pt}

\definecolor{tc0}{RGB}{252,231,220}
\definecolor{tc1}{RGB}{158,167,192}
\definecolor{tc2}{RGB}{21,43,102}
\definecolor{tc3}{RGB}{242,157,109}
\definecolor{tc4}{RGB}{247,198,170}
\definecolor{tc5}{RGB}{1,25,89}
\definecolor{tc6}{RGB}{253,242,236}
\definecolor{tc7}{RGB}{248,201,174}

\noindent
{\color{black}\colorbox{tc0}{\strut \small What}}%
{\color{black}\colorbox{tc1}{\strut \small  is}}%
{\color{white}\colorbox{tc2}{\strut \small  the}}%
{\color{black}\colorbox{tc3}{\strut \small  capital}}%
{\color{black}\colorbox{tc4}{\strut \small  of}}%
{\color{white}\colorbox{tc5}{\strut \small  this}}%
{\color{black}\colorbox{tc6}{\strut \small  country}}%
{\color{black}\colorbox{tc7}{\strut \small ?}}
\\[10pt]
        \midrule
        \textit{(b) Cosine-similarity to final patch vector}\\        
\setlength{\fboxsep}{1pt}

\definecolor{tc0}{RGB}{252,234,224}
\definecolor{tc1}{RGB}{251,227,213}
\definecolor{tc2}{RGB}{254,245,240}
\definecolor{tc3}{RGB}{1,25,89}
\definecolor{tc4}{RGB}{245,177,139}
\definecolor{tc5}{RGB}{248,199,172}
\definecolor{tc6}{RGB}{156,165,190}
\definecolor{tc7}{RGB}{250,220,203}
\definecolor{tc8}{RGB}{247,193,163}
\definecolor{tc9}{RGB}{251,228,215}
\definecolor{tc10}{RGB}{247,196,168}
\definecolor{tc11}{RGB}{248,203,178}
\definecolor{tc12}{RGB}{248,202,177}
\definecolor{tc13}{RGB}{245,181,144}
\definecolor{tc14}{RGB}{242,157,109}

\noindent
{\color{black}\colorbox{tc0}{\strut \small Please}}%
{\color{black}\colorbox{tc1}{\strut \small  give}}%
{\color{black}\colorbox{tc2}{\strut \small  me}}%
{\color{white}\colorbox{tc3}{\strut \small  the}}%
{\color{black}\colorbox{tc4}{\strut \small  sum}}%
{\color{black}\colorbox{tc5}{\strut \small  of}}%
{\color{black}\colorbox{tc6}{\strut \small  the}}%
{\color{black}\colorbox{tc7}{\strut \small  two}}%
{\color{black}\colorbox{tc8}{\strut \small  numbers}}%
{\color{black}\colorbox{tc9}{\strut \small  }}%
{\color{black}\colorbox{tc10}{\strut \small 3}}%
{\color{black}\colorbox{tc11}{\strut \small  and}}%
{\color{black}\colorbox{tc12}{\strut \small  }}%
{\color{black}\colorbox{tc13}{\strut \small 5}}%
{\color{black}\colorbox{tc14}{\strut \small .}}
\\[10pt]
        \midrule
        \textit{(c) KL-Divergence to final patch vector}\\
\setlength{\fboxsep}{1pt}

\definecolor{tc0}{RGB}{242,157,109}
\definecolor{tc1}{RGB}{242,158,110}
\definecolor{tc2}{RGB}{245,183,148}
\definecolor{tc3}{RGB}{246,188,156}
\definecolor{tc4}{RGB}{246,186,152}
\definecolor{tc5}{RGB}{249,207,183}
\definecolor{tc6}{RGB}{245,180,144}
\definecolor{tc7}{RGB}{250,218,200}
\definecolor{tc8}{RGB}{249,210,188}
\definecolor{tc9}{RGB}{250,214,194}

\noindent
{\color{black}\colorbox{tc0}{\strut \small Given}}%
{\color{black}\colorbox{tc1}{\strut \small  a}}%
{\color{black}\colorbox{tc2}{\strut \small  German}}%
{\color{black}\colorbox{tc3}{\strut \small  word}}%
{\color{black}\colorbox{tc5}{\strut \small  respond}}%
{\color{black}\colorbox{tc6}{\strut \small its}}%
{\color{black}\colorbox{tc7}{\strut \small  Chinese}}%
{\color{black}\colorbox{tc8}{\strut \small  translation}}%
{\color{black}\colorbox{tc9}{\strut \small .}}%
\\[10pt]
        \bottomrule
    \end{tabularx}
    

    \caption{
        Visualization of 
        (a) weights attributed by the learned W-MLP to each token activation; 
        (b) Cosine-similarity between the token activations and the compressed patch vector $v$;
        (c) KL-divergence between the token activations and the compressed patch vector $v$.
        A positive metric is depicted in \colorbox{paletteOrange3}{orange}, while a negative value is depicted in \colorbox{paletteBlue3!80}{\textcolor{white}{blue}}
    }
    
    \label{fig:weighting-prompt-qualitative}
\end{table}

\section{Discussion and Future Work}
\label{sec:discussion}

\paragraph{Information Capacity and Accessibility.}
Our experiments suggest that smaller internal representations lead to less accurate compression, independently of the LLM's overall performance. 
This may indicate that compression quality is constrained by the dimensionality of the hidden representation. This also aligns with the idea that sparser representations can be combined more easily, with less destructive interference, as suggested by the superposition perspective of ~\citet{elhage2022superposition}. A preliminary information capacity experiment in~\autoref{app:capacity-experiment} shows that even in the smaller \texttt{Llama3.2-1B} model, one patch vector can encode 9 task families simultaneously without a clear accuracy drop. 
We note that this experiment measures not only the capacity of the vector to encode task information, but also the model's ability to recover and exploit it at inference time.
Nevertheless, the question of the upper bounds on the capacity and accessibility of such representations remains open.

\paragraph{Interpretability by Design.}
Unlike post-hoc explanation methods, the learned weighting mechanism is interpretable by construction: the weights assigned by W-MLP directly expose which tokens the compression has identified as informationally relevant, without requiring any additional analysis pass. 
This property is noteworthy in the context of feature attribution research, where methods such as integrated gradients~\cite{sundararajan2017axiomatic} or black-box attribution approaches~\cite{cai2025gefa} aim to quantify the contribution of each input token to the model output. 
The patching weights offer a structurally similar signal but emerge naturally from the compression objective rather than being retroactively inferred. 
Whether these two families of scores converge empirically is an open question we leave to future work.

\paragraph{Practical Implications.}

Our results point to several possible applications of activation-space compression. 
In consumer API LLMs, system prompts are often reused across many requests and can contribute substantially to the cost of processing the context even after activation caching. Compressing such prompts into a fixed-size activation vector could allow their effect to be reused without re-encoding the full prompt each time. This may be especially attractive for short queries, where the system prompt can dominate the token budget.

In retrieval-augmented generation (RAG), large retrieved text chunks could similarly be compressed once and reused in compact form rather than repeatedly passed to the LLM in full.
This naturally conditions on the method's ability to scale to longer contexts, which we identify as a key future work. 
More speculatively, the same compressed representation could also serve as a semantic vector for similarity search, providing a shared representation for retrieval and generation.

In robotics, vision-language-action (VLA) models~\cite{zitkovich2023rt} condition visuomotor policies on rich instruction contexts that may need to be processed at every control step, making attention overhead a direct latency bottleneck. 
Activation-space compression could potentially alleviate this by encoding complex, reusable instruction or proprioceptive information into a fixed-size representation offline, eliminating sequence-length growth at inference time. 
This mirrors the use of rare tokens as learned capability pointers in fine-tuned VLA systems, but operates in activation rather than embedding space, preserving richer contextual state. 
We leave empirical validation of these applications to future work.

\paragraph{Future Extensions.}
Several directions naturally extend the current work. 
Most immediately, rather than compressing the full prompt into a single patch vector, one could segment the prompt into semantic blocks, each compressed into its own patch token, trading compression ratio for representational fidelity. 
Additionally, enforcing sparsity in individual activations prior to the weighted sum could suppress noise and sharpen the compressed representation.
Finally, with the growing interest in cross-layer transcoders for mechanistic interpretability~\cite{dunefsky2024transcoders}, the aggregation could be extended across layers rather than within a single one, potentially capturing richer hierarchical structure in the compressed activation.

\section{Related Work} \label{sec:related}

\paragraph{Prompt Compression at the Token Level.}
Several works address the redundancy of natural language prompts by compressing them into shorter token sequences. 
Gisting~\cite{mu2023learning} learns to compress prompts into a small number of special tokens, achieving strong results but requiring full LLM fine-tuning. 
AutoCompressors~\cite{chevalier2023adapting} 
and ICAE~\cite{ge2023context} similarly achieve effective long-context compression at the cost of heavy model-specific training. 
KV-cache and prefix caching approaches~\cite{pope2023efficiently} reduce recomputation costs in production systems but operate at the token level and leave the sequence length and attention cost unchanged. 
Our method differs fundamentally: we compress at the \emph{activation} level into a single vector, require no fine-tuning of the target model, and produce a representation that is not human-readable but is directly injectable into the residual stream.

\paragraph{Layer-wise Information and Early Exit.}
Several works demonstrate that transformer layers contain sufficient information to predict the final output well before the last layer~\cite{geva2022transformer, schuster2022confident}. 
Early-exit methods exploit this by halting computation once a confidence threshold is met. 
This body of work directly motivates our injection strategy: patching a compressed activation at an early layer gives the model sufficient depth to process the information before generation, without requiring the full forward pass over the original token sequence.

\paragraph{Activation Steering and Representation Engineering.}
Prior work has shown that high-level behavioral concepts can be encoded in single vectors and injected into specific layers to redirect model behavior~\cite{templeton2024scaling, ardoin2025confabulation}. 
Representation engineering~\cite{zou2023representation} and contrastive activation addition~\cite{panickssery2023steering} similarly manipulate intermediate activations to steer generation. 
Our work is mechanistically related but distinct in objective: prior work steers \emph{behavior}, we compress \emph{prompts}, with the goal of faithful replication rather than redirection.

\paragraph{Superposition and Linearity in Activation Space.}
\citet{elhage2022superposition} provide theoretical grounding for the idea that LLM features are sparsely and superpositionally encoded, implying high redundancy in activation space. 
\citet{liu2023incontext} show empirically that steering vectors can be additively combined while preserving their individual effects. 
These findings motivate our core assumption: that a weighted sum of token activations can preserve the semantic content of the original prompt, and that the resulting vector lies in a region of activation space the model can interpret.

\paragraph{Prompt Redundancy and Information Optimality.}
\citet{melamed2023evil} show that semantically equivalent prompts can differ radically in surface form, suggesting the model's internal task representation is divorced from specific wording. 
\citet{lee2025compress} show that there exists an optimal token count per task, and that additional tokens do not monotonically improve performance. 
Our work pushes this to its logical limit: a single activation vector, raising the question of how much task-relevant information can be packed into the minimal possible representation.

\section{Conclusion} \label{sec:conclusion}

We have shown that relevant prompt information can be compressed into the activation space of an LLM at a controlled accuracy cost. 
A simple learned weighted sum of activations extracted at a middle layer and injected at an early layer proves an effective compression mechanism, as confirmed by our ablations over layer depth and aggregation strategy. 
Beyond its practical implications for inference efficiency, this finding offers new evidence for the linearity of LLM internal representations. 
Meaningful semantic content can be recovered through basic arithmetic over intermediate activations, at a scale beyond what has previously been demonstrated. 
We hope this opens a productive dialogue between efficiency-oriented and mechanistic interpretability research.

\section*{Limitations}

The proposed compression is inherently lossy, and its fidelity under high information-density prompts remains an open question. 
Our evaluation focuses primarily on knowledge retrieval tasks and relatively short prompts. 
Generalization to longer contexts or necessitating reasoning has not been established. Furthermore, the method requires white-box access to the model's internal activations, excluding application to closed-source systems. 
Finally, compression requires a partial forward pass through at least half of the transformer layers, meaning the approach is not cost-free at compression time and is best amortized over repeated reuse of the same context.

\section*{Ethical Considerations}
This work presents a method for compressing prompt activations in LLMs. 
We do not foresee direct ethical implications, though we note that compressed non-human-readable instruction vectors could in principle be used to obscure prompt intent from human reviewers. 
On the positive side, by reducing the per-query computational cost of fixed instruction prompts, the proposed method contributes to more energy-efficient LLM inference, with potential benefits for the environmental footprint of large-scale deployments.

\bibliography{bib}

@inproceedings{ardoin2025confabulation,
    title = {Where Confabulation Lives: Latent Feature Discovery in LLMs},
    author = {Ardoin, Thibaud and Cai, Yi and Wunder, Gerhard},
    booktitle = {Proceedings of the 2025 Conference on Empirical Methods in Natural Language Processing (EMNLP)},
    year = {2025},
    month = aug,
    publisher = {Association for Computational Linguistics},
    url = {}
}

@article{pope2023efficiently,
  title={Efficiently scaling transformer inference},
  author={Pope, Reiner and Douglas, Sholto and Chowdhery, Aakanksha and Devlin, Jacob and Bradbury, James and Heek, Jonathan and Xiao, Kefan and Agrawal, Shivani and Dean, Jeff},
  journal={Proceedings of machine learning and systems},
  volume={5},
  pages={606--624},
  year={2023}
}

@inproceedings{kwon2023efficient,

  title     = {Efficient Memory Management for Large Language Model Serving with PagedAttention},

  author    = {Kwon, Woosuk and Li, Zhuohan and Zhuang, Siyuan and Sheng, Ying and Zheng, Lianmin and Yu, Cody Hao and Gonzalez, Joseph E. and Zhang, Hao and Stoica, Ion},

  booktitle = {Proceedings of the 29th Symposium on Operating Systems Principles},

  pages     = {611--626},

  year      = {2023},

  doi       = {10.1145/3600006.3613165}

}

@inproceedings{sundararajan2017axiomatic,
  title={Axiomatic attribution for deep networks},
  author={Sundararajan, Mukund and Taly, Ankur and Yan, Qiqi},
  booktitle={International conference on machine learning},
  pages={3319--3328},
  year={2017},
  organization={PMLR}
}

@inproceedings{cai2025gefa,
  title={{GEFA}: A general feature attribution framework using proxy gradient estimation},
  author={Cai, Yi and Ardoin, Thibaud and Wunder, Gerhard},
  booktitle={Proceedings of the 42nd International Conference on Machine Learning},
  pages={6165--6192},
  year={2025},
  organization={PMLR}
}

@inproceedings{geva2022transformer,
  title={Transformer feed-forward layers build predictions by promoting concepts in the vocabulary space},
  author={Geva, Mor and Caciularu, Avi and Wang, Kevin and Goldberg, Yoav},
  booktitle={Proceedings of the 2022 conference on empirical methods in natural language processing},
  pages={30--45},
  year={2022}
}

@inproceedings{elhoushi2024layerskip,
  title={Layerskip: Enabling early exit inference and self-speculative decoding},
  author={Elhoushi, Mostafa and Shrivastava, Akshat and Liskovich, Diana and Hosmer, Basil and Wasti, Bram and Lai, Liangzhen and Mahmoud, Anas and Acun, Bilge and Agarwal, Saurabh and Roman, Ahmed and others},
  booktitle={Proceedings of the 62nd Annual Meeting of the Association for Computational Linguistics (Volume 1: Long Papers)},
  pages={12622--12642},
  year={2024}
}

@inproceedings{hendel2023incontext,
  title     = {In-Context Learning Creates Task Vectors},
  author    = {Hendel, Roee and Geva, Mor and Globerson, Amir},
  booktitle = {Findings of the Association for Computational Linguistics: EMNLP 2023},
  pages     = {9318--9333},
  year      = {2023},
  publisher = {Association for Computational Linguistics},
  doi       = {10.18653/v1/2023.findings-emnlp.624}
}

@inproceedings{schuster2022confident,
  title     = {Confident Adaptive Language Modeling},
  author    = {Schuster, Tal and Fisch, Adam and Gupta, Jai and Dehghani, Mostafa and Bahri, Dara and Tran, Vinh Q. and Tay, Yi and Metzler, Donald},
  booktitle = {Advances in Neural Information Processing Systems},
  volume    = {35},
  year      = {2022}
}

@misc{melamed2023evil,
  doi = {10.48550/ARXIV.2311.07064},
  url = {https://arxiv.org/abs/2311.07064},
  author = {Melamed,  Rimon and McCabe,  Lucas H. and Wakhare,  Tanay and Kim,  Yejin and Huang,  H. Howie and Boix-Adsera,  Enric},
  title = {Prompts have evil twins},
  publisher = {arXiv},
  year = {2023},
  copyright = {arXiv.org perpetual,  non-exclusive license}
}

@misc{lee2025compress,
  doi = {10.48550/ARXIV.2503.01141},
  url = {https://arxiv.org/abs/2503.01141},
  author = {Lee,  Ayeong and Che,  Ethan and Peng,  Tianyi},
  title = {How Well do LLMs Compress Their Own Chain-of-Thought? A Token Complexity Approach},
  publisher = {arXiv},
  year = {2025},
  copyright = {arXiv.org perpetual,  non-exclusive license}
}

@inproceedings{chevalier2023adapting,
  title={Adapting language models to compress contexts},
  author={Chevalier, Alexis and Wettig, Alexander and Ajith, Anirudh and Chen, Danqi},
  booktitle={Proceedings of the 2023 Conference on Empirical Methods in Natural Language Processing},
  pages={3829--3846},
  year={2023}
}

@article{ge2023context,
  title={In-context autoencoder for context compression in a large language model},
  author={Ge, Tao and Hu, Jing and Wang, Lei and Wang, Xun and Chen, Si-Qing and Wei, Furu},
  journal={arXiv preprint arXiv:2307.06945},
  year={2023}
}

@article{mu2023learning,
  title={Learning to compress prompts with gist tokens},
  author={Mu, Jesse and Li, Xiang and Goodman, Noah},
  journal={Advances in Neural Information Processing Systems},
  volume={36},
  pages={19327--19352},
  year={2023}
}

@article{liu2023incontext,
  author    = {Sheng Liu and Haotian Ye and Lei Xing and James Zou},
  title     = {In-Context Vectors: Making In-Context Learning More Effective
               and Controllable Through Latent Space Steering},
  journal   = {arXiv preprint arXiv:2311.06668},
  year      = {2023},
}

@inproceedings{subramani2022extracting,
  title={Extracting latent steering vectors from pretrained language models},
  author={Subramani, Nishant and Suresh, Nivedita and Peters, Matthew E},
  booktitle={Findings of the Association for Computational Linguistics: ACL 2022},
  pages={566--581},
  year={2022}
}

@article{panickssery2023steering,
  author    = {Nina Panickssery and Nick Gabrieli and Julian Schulz and
               Meg Tong and Evan Hubinger and Alexander Matt Turner},
  title     = {Steering {L}lama 2 via Contrastive Activation Addition},
  journal   = {arXiv preprint arXiv:2312.06681},
  year      = {2023},
}

@article{zou2023representation,
  author    = {Andy Zou and Long Phan and Sarah Chen and James Campbell and
               Phillip Guo and Richard Ren and Alexander Pan and
               Xuwang Yin and Mantas Mazeika and Ann-Kathrin Dombrowski and
               Shashwat Goel and Nathaniel Li and Michael J. Byun and
               Zifan Wang and Alex Mallen and Steven Basart and
               Sanmi Koyejo and Dawn Song and Matt Fredrikson and others},
  title     = {Representation Engineering: A Top-Down Approach to {AI} Transparency},
  journal   = {arXiv preprint arXiv:2310.01405},
  year      = {2023},
}

@article{Marks2023,
  doi = {10.48550/ARXIV.2310.06824},
  url = {https://arxiv.org/abs/2310.06824},
  author = {Marks,  Samuel and Tegmark,  Max},
  title = {The Geometry of Truth: Emergent Linear Structure in Large Language Model Representations of True/False Datasets},
  journal = {arXiv preprint arXiv:2310.06824},
  publisher = {arXiv},
  year = {2023},
  copyright = {Creative Commons Attribution 4.0 International}
}

@inproceedings{chauhan2026punctuations,
  title={Punctuations and Predicates in Language Models},
  author={Chauhan, Sonakshi and Chaudhary, Maheep and Kiu, Choy Kwan and Nellessen, Samuel and Schoots, Nandi},
  booktitle={Findings of the Association for Computational Linguistics: EACL 2026},
  pages={5622--5636},
  year={2026}
}

@inproceedings{xiao2024efficient,
  title={Efficient streaming language models with attention sinks},
  author={Xiao, Guangxuan and Tian, Yuandong and Chen, Beidi and Han, Song and Lewis, Mike},
  booktitle={International Conference on Learning Representations},
  volume={2024},
  pages={21875--21895},
  year={2024}
}

@misc{templeton2024scaling,
  title   = {Scaling Monosemanticity: Extracting Interpretable Features from Claude 3 Sonnet},
  author  = {Templeton, Adly and Conerly, Tom and Marcus, Jonathan and Lindsey, Jack and Bricken, Trenton and Chen, Brian and Pearce, Adam and Citro, Craig and Ameisen, Emmanuel and Jones, Andy and Cunningham, Hoagy and Turner, Nicholas L. and McDougall, Callum and MacDiarmid, Monte and Freeman, C. Daniel and Sumers, Theodore R. and Rees, Edward and Batson, Joshua and Jermyn, Adam and Carter, Shan and Olah, Chris and Henighan, Tom},
  year    = {2024},
  howpublished = {\url{https://transformer-circuits.pub/2024/scaling-monosemanticity/}}
}

@misc{elhage2022superposition,
  title        = {Toy Models of Superposition},
  author       = {Elhage, Nelson and Hume, Tristan and Olsson, Catherine and Schiefer, Nicholas and Henighan, Tom and Kravec, Shauna and Hatfield-Dodds, Zac and Lasenby, Robert and Drain, Dawn and Chen, Carol and Grosse, Roger and McCandlish, Sam and Kaplan, Jared and Amodei, Dario and Wattenberg, Martin and Olah, Christopher},
  year         = {2022},
  howpublished = {\url{https://transformer-circuits.pub/2022/toy_model/}}
}

@article{dunefsky2024transcoders,
  title={Transcoders find interpretable llm feature circuits},
  author={Dunefsky, Jacob and Chlenski, Philippe and Nanda, Neel},
  journal={Advances in Neural Information Processing Systems},
  volume={37},
  pages={24375--24410},
  year={2024}
}

@inproceedings{Mikolov2013linguistic,
    title = "Linguistic Regularities in Continuous Space Word Representations",
    author = "Mikolov, Tomas  and
      Yih, Wen-tau  and
      Zweig, Geoffrey",
    editor = "Vanderwende, Lucy  and
      Daum{\'e} III, Hal  and
      Kirchhoff, Katrin",
    booktitle = "Proceedings of the 2013 Conference of the North {A}merican Chapter of the Association for Computational Linguistics: Human Language Technologies",
    month = jun,
    year = "2013",
    address = "Atlanta, Georgia",
    publisher = "Association for Computational Linguistics",
    url = "https://aclanthology.org/N13-1090",
    pages = "746--751",
}

@article{liu2026ministral,
  title={Ministral 3},
  author={Liu, Alexander H and Khandelwal, Kartik and Subramanian, Sandeep and Jouault, Victor and Rastogi, Abhinav and Sad{\'e}, Adrien and Jeffares, Alan and Jiang, Albert and Cahill, Alexandre and Gavaudan, Alexandre and others},
  journal={arXiv preprint arXiv:2601.08584},
  year={2026}
}

@misc{qwen3.5,
    title  = {{Qwen3.5}: Towards Native Multimodal Agents},
    author = {{Qwen Team}},
    month  = {February},
    year   = {2026},
    url    = {https://qwen.ai/blog?id=qwen3.5}
}

@article{Grattafiori2024,
  doi = {10.48550/ARXIV.2407.21783},
  url = {https://arxiv.org/abs/2407.21783},
  author = {Grattafiori,  Aaron and Dubey,  Abhimanyu and Jauhri,  Abhinav and Pandey,  Abhinav and Kadian,  Abhishek and Al-Dahle,  Ahmad and Letman,  Aiesha and Mathur,  Akhil and Schelten,  Alan and Vaughan,  Alex and Yang,  Amy and Fan,  Angela and Goyal,  Anirudh and Hartshorn,  Anthony and Yang,  Aobo and Mitra,  Archi and Sravankumar,  Archie},
  title = {The Llama 3 Herd of Models},
  journal = {arXiv preprint arXiv:2407.21783},
  publisher = {arXiv},
  year = {2024},
  copyright = {arXiv.org perpetual,  non-exclusive license}
}

@inproceedings{zitkovich2023rt,
  title={Rt-2: Vision-language-action models transfer web knowledge to robotic control},
  author={Zitkovich, Brianna and Yu, Tianhe and Xu, Sichun and Xu, Peng and Xiao, Ted and Xia, Fei and Wu, Jialin and Wohlhart, Paul and Welker, Stefan and Wahid, Ayzaan and others},
  booktitle={Conference on Robot Learning},
  pages={2165--2183},
  year={2023},
  organization={PMLR}
}

@article{hendrycks2020measuring,
  title={Measuring massive multitask language understanding},
  author={Hendrycks, Dan and Burns, Collin and Basart, Steven and Zou, Andy and Mazeika, Mantas and Song, Dawn and Steinhardt, Jacob},
  journal={arXiv preprint arXiv:2009.03300},
  year={2020}
}

@article{clark2018think,
  title={Think you have solved question answering? try arc, the ai2 reasoning challenge},
  author={Clark, Peter and Cowhey, Isaac and Etzioni, Oren and Khot, Tushar and Sabharwal, Ashish and Schoenick, Carissa and Tafjord, Oyvind},
  journal={arXiv preprint arXiv:1803.05457},
  year={2018}
}

\clearpage

\appendix

\section{Hand-made Weighting Method}
\label{app:hand-made-weights}

As a simple proof-of-concept baseline, we construct a patch vector directly from the prompt activations using a hand-designed weighting rule. This experiment tests whether task-relevant information can be recovered from the activation sequence of a task prompt using a simple hand-crafted heuristic without training an additional model.

\paragraph{Setup.}
We consider the country-capital task and use the base Llama-3.1-8B model in a completion-style setting. The source prompt is
$$
\texttt{Give me the capital city of a country I name:}
$$
and we extract the hidden states of all prompt tokens at an intermediate layer. At evaluation time, queries are written in the form
$$
\texttt{:Country:}
$$
For example, as \texttt{:France:}. The first colon acts as a placeholder token and is the position at which the patch is injected.

\paragraph{Hand-made Vector Construction.}
Let 
$$
p=(t_1, \dots, t_T)$$
denote the tokenized source prompt, and let 
$$h_i^{(m)} \in \mathbb{R}^d$$
denote the hidden state of token $t_i$ at the extraction layer $m$. In our experiment, we use extraction layer $m=12.$

The hand-made vector consists of two components. 
First, we compute a position-weighted sum of all prompt activations:
$$
v_{\mathrm{pos}}= \sum_{i=1}^T (0.1i)h_i^{(m)}.
$$
This gives larger weights to later tokens in the prompt. The heuristic is that later activations may contain a more complete representation of the task instruction.
Second, we manually amplify a small set of task-relevant token positions. For the prompt above, we use 
$$
S=\{4,5,T\},$$
which corresponds to the tokens associated with \texttt{capital}, \texttt{city}, and the final token \texttt{:}. We define
$$
v_{\mathrm{key}}= \alpha \sum_{i \in S} h_i^{(m)}$$
with amplification factor $\alpha=6.$ The final patch vector is
$$
v= v_{\mathrm{pos}} + v_{\mathrm{key}}, 
$$
which is normalized before being injected:

$$
\hat{v}=\frac{v}{\|v\|_2}.
$$

\paragraph{Patching Procedure.}
For a query of the form \texttt{:Country:}, the first colon is used as the placeholder position. In the base completion model, this corresponds to token position $1$, after the beginning-of-sequence token:

$$
\texttt{<bos> : Country :}.
$$
We inject the patch vector at layer $e=2$ by replacing the hidden state of the placeholder token:
$$
h_{\mathrm{placeholder}}^{(e)} \leftarrow \lambda \hat v,
$$
where $\lambda$ is a scalar amplification factor. In the reproduced experiment, we use $\lambda=100{,}000$. This is replacement patching rather than additive patching. The full configuration is summarized in \autoref{tab:handmade-config}.

\begin{table}[htbp]
\centering
\begin{tabular}{ll}
\toprule
Component & Value \\
\midrule
Model & Llama-3.1-8B base \\
Prompt format & Completion-style \\
Source prompt & \texttt{Give me the capital city} \\
&\texttt{of a country I name:} \\
Query format & \texttt{:Country:} \\
Extraction layer $m$ & 12 \\
Patching layer $e$ & 2 \\
Selected positions $S$ & $\{4,5,T\}$ \\
Amplification $\alpha$ & 6 \\
Scalar $\lambda$ & $100{,}000$ \\
Patch operation & Replacement \\
Accuracy & $85.1\%$ \\
\bottomrule
\end{tabular}
\caption{Configuration of the hand-made weighting experiment.}
\label{tab:handmade-config}
\end{table}

\paragraph{Results and Ablation.}
With the setup in ~\autoref{tab:handmade-config}, the hand-made patch vector achieves $85.1\%$ accuracy on the country-capital training split for the base Llama-3.1-8B model.

To understand the contribution of the two terms, we also evaluate them separately. The results are shown in ~\autoref{tab:handmade-ablation}. The position-weighted component alone reaches $56.4\%$, showing that a rough positional weighting already captures some task information. The manually selected token component is stronger, reaching $80.2\%$.  Combining both terms gives the best result.

\begin{table}[htbp]
\centering
\begin{tabular}{lc}
\toprule
Patch vector & Accuracy \\
\midrule
$v_{\mathrm{pos}}$ only & $56.4\%$ \\
$v_{\mathrm{key}}$ only & $80.2\%$ \\
$v_{\mathrm{pos}} + v_{\mathrm{key}}$ & $85.1\%$ \\
\bottomrule
\end{tabular}
\caption{Ablation of the hand-made patch vector on the country-capital training split.}
\label{tab:handmade-ablation}
\end{table}

This ablation suggests that task-relevant information is not distributed uniformly across the prompt activation sequence. A small number of semantically important token activations already carry much of the useful signal. At the same time, choosing these tokens by hand is delicate. Namely, the construction depends on the prompt wording, tokenization, patch position, and scaling factor.

This motivates the learned W-MLP aggregation model. Rather than fixing the weights $w_i^{\mathrm{hand}}$ manually, W-MLP learns the token weights from data and therefore can adapt the aggregation to different prompts and tasks. In particular, the hand-made construction does not transfer directly to the chat-formatted Llama-3.1-8B-Instruct model, where performance remained close to $5.9\%$ across patching layers. We thus treat this hand-made vector as a proof of concept rather than a robust general-purpose model.

\section{Datasets and Prompt Formats}
\label{app:datasets}
We use two types of datasets in our experiments: a collection of synthetic dictionary-style toy tasks and the ARC-Easy multiple-choice question answering benchmark. We now describe the input-output format, prompt format, and evaluation criterion for both settings.

\paragraph{Toy Task Dataset}
The toy task dataset consists of simple mapping tasks. Each example is a pair $(x,y)$, where $x$ is an input entity and $y$ is the target string. The task prompt specifies which mapping should be applied. For example, in the country-capital task, the input is a country name and the target is its capital city.

In the main in-distribution (ID) experiments, we use eight toy tasks: capitals, ISO country codes, continents, event years, English-to-French translation, art authorship, antonyms, and currency codes. We additionally evaluate out-of-distribution (OOD) transfer on French-to-English translation, chemical element symbols, and addition of small integers. Representative examples are shown in \autoref{tab:toy-task-examples}.

\begin{table}[h!]
    \centering
    \small
    \begin{tabularx}{\columnwidth}{llXX}
         \toprule
         Split & Task & Input $x$ & Target $y$  \\
         \midrule
         ID & Country-capital & France & Paris \\
         ID & ISO code & France & FR\\
         ID & Continent & France & Europe \\
         ID & Event year & French revolution & 1789 \\
         ID & En$\to$Fr & dog & chien  \\
         ID & Art author & Mona Lisa & Leonardo da Vinci \\
         ID & Antonym & hot & cold \\
         ID & Currency & Japan & yen \\
         OOD & Fr$\to$En & chien & dog \\
         OOD & Element symbol & Hydrogen & H \\
         OOD & Addition & $17+25$ & 42 \\
         \bottomrule
    \end{tabularx}
    \caption{Examples of toy task input-output pairs.}
    \label{tab:toy-task-examples}
\end{table}

For each toy task, we use a set of approximately $10$ different natural-language prompts that describe the mapping to be performed. These prompts are used to extract the activation sequence from which the patch vector is constructed. Examples are given in~\autoref{tab:toy-prompt-examples}.

\begin{table}[h!]
    \centering
    \small
    \begin{tabularx}{\columnwidth}{lX}
        \toprule
        Task & Example prompt \\
        \midrule
        Country-capital & \texttt{Give me the capital city of a country I name} \\
        ISO country code & \texttt{Return the ISO code of the country I name} \\
        Continent & \texttt{Return the continent of the country I name} \\
        Event year & \texttt{Return the year of the event I name} \\
        English-French & \texttt{Translate the English word to French} \\
        Art author & \texttt{Return the author of the work I name} \\
        Antonym & \texttt{Give me the antonym of the word I provide} \\
        Currency & \texttt{Return the currency of the country I name} \\
        \bottomrule
    \end{tabularx}
    \caption{Example prompts for the toy task dataset.}
    \label{tab:toy-prompt-examples}
\end{table}

At evaluation time, the model receives the input entity $x$ together with the patch vector. The generated output is marked correct if the target string $y$ occurs in the generated continuation after lowercasing. This string-based criterion is used because the target may consist of more than one token.

\paragraph{Toy Task Prompt Format.}
For the instruct-model experiments, the task prompt is encoded into the patch vector, while the query contains only a placeholder and the input entity. In the notation of the main text, the compressed task prompt is the information that is replaced by the patch vector. Abstractly, the injection prompt has the form
$$
\begin{aligned}
\texttt{[PATCH] } x,
\end{aligned}
$$
where $x$ is the input entity. For example, for the country-capital task and input \texttt{Italy}, the intended behavior is
$$
\texttt{[PATCH] Italy}
\quad \longrightarrow \quad
\texttt{Rome}.
$$

In the actual Llama-3.1-8B-Instruct implementation, prompts are formatted using the model's chat template. The placeholder is placed in the system and the input entity is placed in the user message:

$$
\begin{aligned}
    \texttt{system:} &\quad \texttt{[PATCH]} \\
    \texttt{user:} &\quad x.
\end{aligned}
$$

After applying the chat template, the placeholder appears at a fixed token position in the system message. In our Llama-3.1-8B-Instruct, this was token position $6$, and the patch vector is injected at that position.

\paragraph{ARC-Easy}
In addition to the toy task dataset, we also evaluate on ARC-Easy, the Easy split of the AI2 Reasoning Challenge~\cite{clark2018think}. ARC-Easy is a multiple-choice science question answering benchmark. Each example consists of a natural-language question, four answer letter candidates, and a target answer letter. In our compression setup, the question is encoded into the patch vector, while the model receives the answer choices. Thus, if compression fails, the model has access to the answer choices but not to the question. An example is given in~\autoref{tab:arc-format}.

\begin{table}[htbp]
\centering
\begin{tabular}{ll}
\toprule
Field & Example \\
\midrule
Question & \texttt{Which object is attracted}  \\
& \texttt{to a magnet?}\\
Choice A & \texttt{wooden spoon} \\
Choice B & \texttt{iron nail} \\
Choice C & \texttt{plastic cup} \\
Choice D & \texttt{glass bottle} \\
Target & \texttt{B} \\
\bottomrule
\end{tabular}
\caption{Example format for ARC-Easy.}
\label{tab:arc-format}
\end{table}

For ARC-Easy, the extraction prompt contains the question. The injection prompt contains the placeholder and the answer of the choices, together with an instruction to answer with only the letter of the correct option. Abstractly, the format is
$$
\begin{aligned}
    \text{Extraction:} \quad & \texttt{Question} \\
    \text{Injection:} \quad & \texttt{Reply with ONLY the letter of} \\
    & \texttt{the correct answer} \\
                            & \texttt{[PATCH] (A) ... (B) ...}\\
                            & \texttt{(C) ... (D) ...}
\end{aligned}
$$

\section{Multi-Task Encoding in a Single Vector}
\label{app:capacity-experiment}

To probe the information capacity of a single patched activation, we design a complementary experiment in which we directly optimize a single vector for downstream task performance, rather than a compression network. 
Concretely, for a set of $k$ tasks, we train a single patch vector $\mathbf{v}$ with backpropagation such that, when injected into the model at the standard patching layer, the model successfully performs all $k$ tasks on the given input entities. 
No prompt is provided at inference time: the vector alone must encode the full multi-task instruction. In order to recover the correct task the model is prompted with the id of the task:

$$
\begin{aligned}
    \texttt{system:} &\quad \texttt{[PATCH]} \\
    \texttt{user:} &\quad \texttt{Task i: [ENTITY]} 
\end{aligned}
$$

\begin{figure}[h!]
    \centering
    \includegraphics[width=0.86\columnwidth]{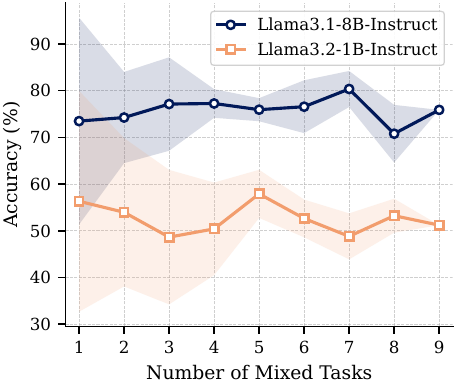}
    \caption{
    Multi-task encoding capacity of a single optimized patch vector. For each task group size $k$, we optimize one vector over a random subset of $k$ tasks. Results are averaged over 5 runs.
    }
    \label{fig:multitask-capacity}
\end{figure}

The results are shown in \autoref{fig:multitask-capacity}.
\vspace{1em}

We train this vector for an increasing number of tasks, monitoring whether combining all tasks into a single fixed representation causes a collapse in accuracy. 
Surprisingly, we find no such collapse within our experimental range with up to 9 different tasks: 
a single vector of dimension $4096$ for \texttt{Llama3.1-8B-Instruct}, and $1024$ for the \texttt{Llama3.2-1B-Instruct} model, remains sufficient to encode all tested tasks simultaneously without measurable degradation. 
This suggests that the activation space of these models has substantially higher information capacity than a single task requires. This potentially also implies that task instructions are not competing for the same representational directions.

We note that the precise capacity limit of a single patched vector remains an open question. Establishing this limit rigorously would require a broader and more diverse task set, and we leave this to future work.

\section{Examples of Weight Attribution and Similarity Measures}
\label{app:weighting-examples}
We provide a qualitative visualization of the learned W-MLP weights and their relation to the original prompt-token activations. These examples complement the hand-made weighting experiment in~\autoref{app:hand-made-weights}. The hand-made construction suggested that a small number of semantically important prompt tokens can carry a large part of the useful task information. The visualizations below show that the learned W-MLP weights follow a similar pattern. 

\paragraph{Metrics.}
Let
$$
H^{(m)}= (h^{(m)}_1, \dots, h^{(m)}_T)
$$
be the token activation sequence of a prompt $p$ at layer $m$, and let $v$ be the learned patch produced by the W-MLP. For each token position $i$, we compute the cosine similarity

$$
\operatorname{sim}(v,h^{(m)}_i) = \frac{\langle v, h^{(m)}_i \rangle}{\|v\|_2 \, \|h^{(m)}_i\|_2}.
$$
This measures how close the final patch vector is to each original prompt-token activation in representation space.

Next to cosine similarity, we compute a KL-divergence-based comparison between the output distribution induced by the patch vector and the output distribution induced by the corresponding prompt-token activation. Concretely, for each token activation $h_i^{(m)}$, we compare the output distribution induced by $h_i^{(m)}$ with the output distribution induced by the patch vector $v$:

$$
D_{\mathrm{KL}}
    \bigl(
        p_v(\cdot \mid x)
        \,\|\, 
        p_i(\cdot \mid x)
    \bigr).
$$

Lower values indicate that the token activation induces a predictive behavior closer to that of the final patch vector.

Finally, we visualize the scalar weights predicted by the W-MLP:
\vspace{-0.3em}
$$
    w_i = M_\theta(h_i^{(m)}),
    \qquad
    v = \sum_{i=1}^{T} w_i h_i^{(m)}.
$$

These weights indicate how strongly each prompt-token activation contributes to the final compressed representation.

\begin{figure}[h!]
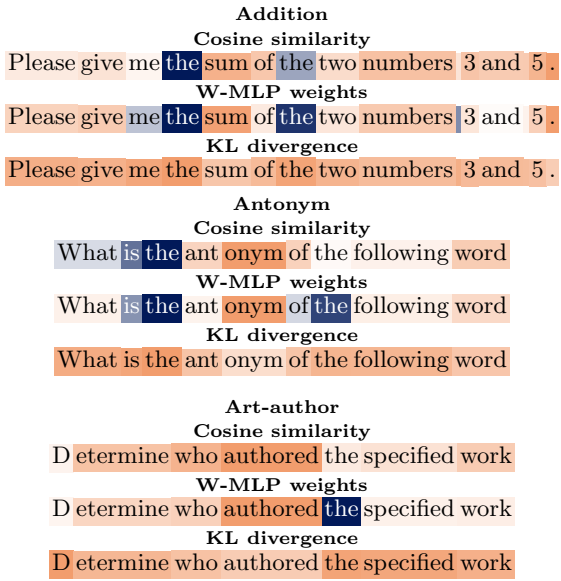
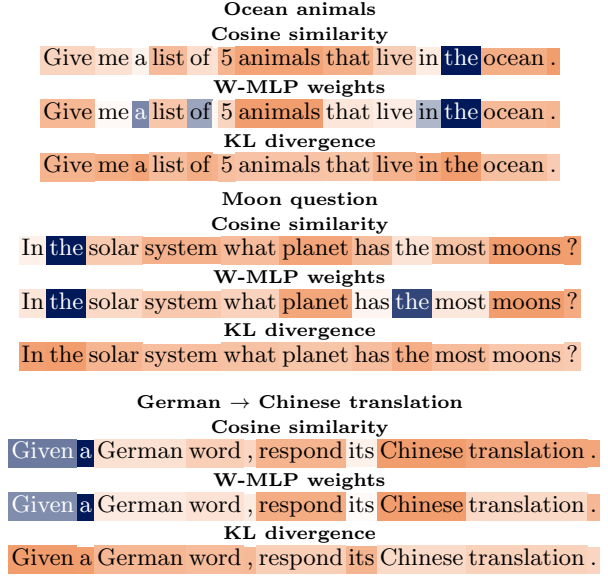

\centering
\begin{minipage}[t]{\columnwidth}
\centering
\scriptsize
\vspace{0.4em}
\textbf{Addition}\par
\vspace{0.15em}

\textbf{Cosine similarity}\par
\input{figures/weighted_texts/token_weights.addition_3_and_5.similarities.tex}\par

\vspace{0.15em}
\textbf{W-MLP weights}\par
\input{figures/weighted_texts/token_weights.addition_3_and_5.weights.tex}\par

\vspace{0.15em}
\textbf{KL divergence}\par
\input{figures/weighted_texts/token_weights.addition_3_and_5.kl_divergences.tex}\par

\end{minipage}%

\vspace{0.5em}
\begin{minipage}[t]{\columnwidth}
\centering
\scriptsize
\vspace{0.4em}
\textbf{Antonym}\par
\vspace{0.15em}

\textbf{Cosine similarity}\par
\input{figures/weighted_texts/token_weights.antonyme.similarities.tex}\par

\vspace{0.15em}
\textbf{W-MLP weights}\par
\input{figures/weighted_texts/token_weights.antonyme.weights.tex}\par

\vspace{0.15em}
\textbf{KL divergence}\par
\input{figures/weighted_texts/token_weights.antonyme.kl_divergences.tex}\par

\end{minipage}%

\vspace{0.5em}
\begin{minipage}[t]{\columnwidth}
\centering
\scriptsize
\vspace{0.4em}
\textbf{Art-author}\par
\vspace{0.15em}

\textbf{Cosine similarity}\par
\input{figures/weighted_texts/token_weights.art_author.similarities.tex}\par

\vspace{0.15em}
\textbf{W-MLP weights}\par
\input{figures/weighted_texts/token_weights.art_author.weights.tex}\par

\vspace{0.15em}
\textbf{KL divergence}\par
\input{figures/weighted_texts/token_weights.art_author.kl_divergences.tex}\par

\end{minipage}%

\vspace{-0.9em}
\caption{Qualitative token-level visualizations.}
\label{fig:token-weights-examples-2}
\end{figure}

\begin{figure*}[htbp]
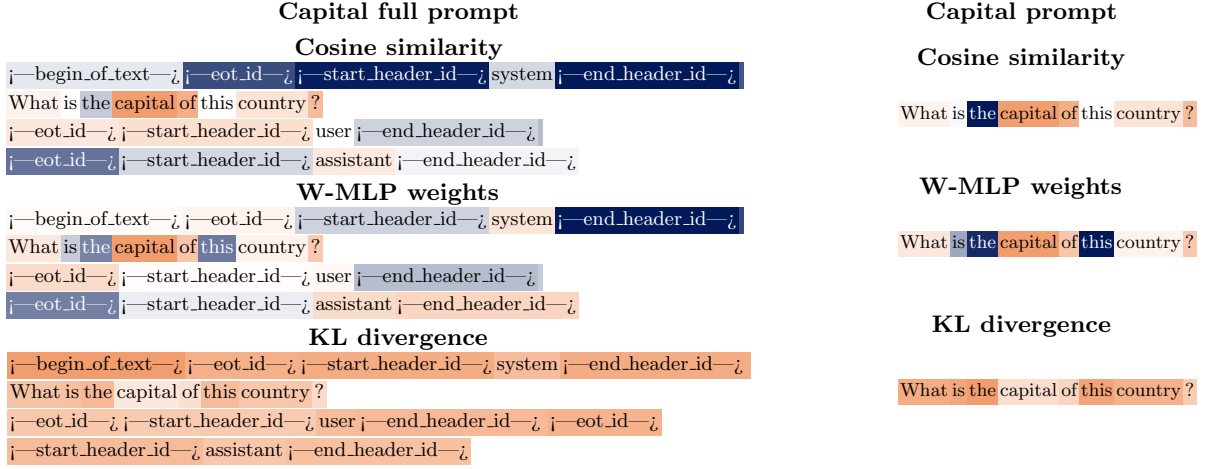

\centering

\begin{minipage}[t]{0.67\textwidth}
\centering
\footnotesize
\textbf{Capital full prompt}\par
\vspace{0.4em}

\textbf{Cosine similarity}
\hspace*{-2.7cm}%
\resizebox{1.12\linewidth}{!}{%
\setlength{\fboxsep}{1pt}

\definecolor{tc0}{RGB}{229,232,238}
\definecolor{tc1}{RGB}{58,77,126}
\definecolor{tc2}{RGB}{6,30,93}
\definecolor{tc3}{RGB}{212,216,227}
\definecolor{tc4}{RGB}{1,25,89}
\definecolor{tc5}{RGB}{68,85,133}
\definecolor{tc6}{RGB}{253,242,236}
\definecolor{tc7}{RGB}{254,254,254}
\definecolor{tc8}{RGB}{195,201,216}
\definecolor{tc9}{RGB}{242,157,109}
\definecolor{tc10}{RGB}{244,175,136}
\definecolor{tc11}{RGB}{255,254,253}
\definecolor{tc12}{RGB}{251,227,213}
\definecolor{tc13}{RGB}{246,188,155}
\definecolor{tc14}{RGB}{252,231,220}
\definecolor{tc15}{RGB}{251,223,207}
\definecolor{tc16}{RGB}{255,255,254}
\definecolor{tc17}{RGB}{206,210,223}
\definecolor{tc18}{RGB}{180,188,206}
\definecolor{tc19}{RGB}{101,115,154}
\definecolor{tc20}{RGB}{210,215,226}
\definecolor{tc21}{RGB}{252,234,224}
\definecolor{tc22}{RGB}{243,244,247}

\begin{tabular}{@{}l@{}}
\noindent
{\color{black}\colorbox{tc0}{\strut \small <|begin\_of\_text|>}}%
{\color{white}\colorbox{tc1}{\strut \small <|eot\_id|>}}%
{\color{white}\colorbox{tc2}{\strut \small <|start\_header\_id|>}}%
{\color{black}\colorbox{tc3}{\strut \small system}}%
{\color{white}\colorbox{tc4}{\strut \small <|end\_header\_id|>}}%
{\color{white}\colorbox{tc5}{\strut \small 
}}%
\\[0.2em]
{\color{black}\colorbox{tc6}{\strut \small What}}%
{\color{black}\colorbox{tc7}{\strut \small  is}}%
{\color{black}\colorbox{tc8}{\strut \small  the}}%
{\color{black}\colorbox{tc9}{\strut \small  capital}}%
{\color{black}\colorbox{tc10}{\strut \small  of}}%
{\color{black}\colorbox{tc11}{\strut \small  this}}%
{\color{black}\colorbox{tc12}{\strut \small  country}}%
{\color{black}\colorbox{tc13}{\strut \small ?}}%
\\[0.2em]
{\color{black}\colorbox{tc14}{\strut \small <|eot\_id|>}}%
{\color{black}\colorbox{tc15}{\strut \small <|start\_header\_id|>}}%
{\color{black}\colorbox{tc16}{\strut \small user}}%
{\color{black}\colorbox{tc17}{\strut \small <|end\_header\_id|>}}%
{\color{black}\colorbox{tc18}{\strut \small 
}}%
\\[0.2em]
{\color{white}\colorbox{tc19}{\strut \small <|eot\_id|>}}%
{\color{black}\colorbox{tc20}{\strut \small <|start\_header\_id|>}}%
{\color{black}\colorbox{tc21}{\strut \small assistant}}%
{\color{black}\colorbox{tc22}{\strut \small <|end\_header\_id|>}}%
\end{tabular}
}

\vspace{0.25em}
\textbf{W-MLP weights}
\hspace*{-2.7cm}%
\resizebox{1.12\linewidth}{!}{%
\setlength{\fboxsep}{1pt}

\definecolor{tc0}{RGB}{254,250,247}
\definecolor{tc1}{RGB}{254,244,239}
\definecolor{tc2}{RGB}{205,210,222}
\definecolor{tc3}{RGB}{251,229,216}
\definecolor{tc4}{RGB}{1,25,89}
\definecolor{tc5}{RGB}{52,72,123}
\definecolor{tc6}{RGB}{252,231,220}
\definecolor{tc7}{RGB}{199,204,218}
\definecolor{tc8}{RGB}{119,132,166}
\definecolor{tc9}{RGB}{242,157,109}
\definecolor{tc10}{RGB}{247,198,170}
\definecolor{tc11}{RGB}{107,121,158}
\definecolor{tc12}{RGB}{253,242,236}
\definecolor{tc13}{RGB}{248,201,174}
\definecolor{tc14}{RGB}{250,219,201}
\definecolor{tc15}{RGB}{255,252,251}
\definecolor{tc16}{RGB}{254,247,243}
\definecolor{tc17}{RGB}{182,189,207}
\definecolor{tc18}{RGB}{197,203,217}
\definecolor{tc19}{RGB}{115,128,164}
\definecolor{tc20}{RGB}{235,237,242}
\definecolor{tc21}{RGB}{250,220,202}
\definecolor{tc22}{RGB}{250,214,194}

\begin{tabular}{@{}l@{}}
\noindent
{\color{black}\colorbox{tc0}{\strut \small <|begin\_of\_text|>}}%
{\color{black}\colorbox{tc1}{\strut \small <|eot\_id|>}}%
{\color{black}\colorbox{tc2}{\strut \small <|start\_header\_id|>}}%
{\color{black}\colorbox{tc3}{\strut \small system}}%
{\color{white}\colorbox{tc4}{\strut \small <|end\_header\_id|>}}%
{\color{white}\colorbox{tc5}{\strut \small 
}}%
\\[0.2em]
{\color{black}\colorbox{tc6}{\strut \small What}}%
{\color{black}\colorbox{tc7}{\strut \small  is}}%
{\color{white}\colorbox{tc8}{\strut \small  the}}%
{\color{black}\colorbox{tc9}{\strut \small  capital}}%
{\color{black}\colorbox{tc10}{\strut \small  of}}%
{\color{white}\colorbox{tc11}{\strut \small  this}}%
{\color{black}\colorbox{tc12}{\strut \small  country}}%
{\color{black}\colorbox{tc13}{\strut \small ?}}%
\\[0.2em]
{\color{black}\colorbox{tc14}{\strut \small <|eot\_id|>}}%
{\color{black}\colorbox{tc15}{\strut \small <|start\_header\_id|>}}%
{\color{black}\colorbox{tc16}{\strut \small user}}%
{\color{black}\colorbox{tc17}{\strut \small <|end\_header\_id|>}}%
{\color{black}\colorbox{tc18}{\strut \small 
}}%
\\[0.2em]
{\color{white}\colorbox{tc19}{\strut \small <|eot\_id|>}}%
{\color{black}\colorbox{tc20}{\strut \small <|start\_header\_id|>}}%
{\color{black}\colorbox{tc21}{\strut \small assistant}}%
{\color{black}\colorbox{tc22}{\strut \small <|end\_header\_id|>}}%
\end{tabular}
}

\vspace{0.25em}
\textbf{KL divergence}
\hspace*{-2.7cm}%
\resizebox{1.12\linewidth}{!}{%
\setlength{\fboxsep}{1pt}

\definecolor{tc0}{RGB}{242,157,109}
\definecolor{tc1}{RGB}{245,176,137}
\definecolor{tc2}{RGB}{244,172,131}
\definecolor{tc3}{RGB}{246,188,155}
\definecolor{tc4}{RGB}{244,172,132}
\definecolor{tc5}{RGB}{244,174,134}
\definecolor{tc6}{RGB}{247,195,166}
\definecolor{tc7}{RGB}{247,192,161}
\definecolor{tc8}{RGB}{246,184,149}
\definecolor{tc9}{RGB}{252,231,220}
\definecolor{tc10}{RGB}{251,224,209}
\definecolor{tc11}{RGB}{246,188,156}
\definecolor{tc12}{RGB}{248,199,171}
\definecolor{tc13}{RGB}{250,219,201}
\definecolor{tc14}{RGB}{248,199,172}
\definecolor{tc15}{RGB}{248,203,177}
\definecolor{tc16}{RGB}{246,188,155}
\definecolor{tc17}{RGB}{246,184,150}
\definecolor{tc18}{RGB}{246,185,150}
\definecolor{tc19}{RGB}{245,178,140}
\definecolor{tc20}{RGB}{246,187,154}
\definecolor{tc21}{RGB}{248,199,171}
\definecolor{tc22}{RGB}{247,192,161}

\begin{tabular}{@{}l@{}}
\noindent
{\color{black}\colorbox{tc0}{\strut \small <|begin\_of\_text|>}}%
{\color{black}\colorbox{tc1}{\strut \small <|eot\_id|>}}%
{\color{black}\colorbox{tc2}{\strut \small <|start\_header\_id|>}}%
{\color{black}\colorbox{tc3}{\strut \small system}}%
{\color{black}\colorbox{tc4}{\strut \small <|end\_header\_id|>}}%
{\color{black}\colorbox{tc5}{\strut \small 
}}%
\\[0.2em]
{\color{black}\colorbox{tc6}{\strut \small What}}%
{\color{black}\colorbox{tc7}{\strut \small  is}}%
{\color{black}\colorbox{tc8}{\strut \small  the}}%
{\color{black}\colorbox{tc9}{\strut \small  capital}}%
{\color{black}\colorbox{tc10}{\strut \small  of}}%
{\color{black}\colorbox{tc11}{\strut \small  this}}%
{\color{black}\colorbox{tc12}{\strut \small  country}}%
{\color{black}\colorbox{tc13}{\strut \small ?}}
\\[0.2em]
{\color{black}\colorbox{tc14}{\strut \small <|eot\_id|>}}%
{\color{black}\colorbox{tc15}{\strut \small <|start\_header\_id|>}}%
{\color{black}\colorbox{tc16}{\strut \small user}}%
{\color{black}\colorbox{tc17}{\strut \small <|end\_header\_id|>}}%
{\color{black}\colorbox{tc18}{\strut \small 
}}%
{\color{black}\colorbox{tc19}{\strut \small <|eot\_id|>}}%
\\[0.2em]
{\color{black}\colorbox{tc20}{\strut \small <|start\_header\_id|>}}%
{\color{black}\colorbox{tc21}{\strut \small assistant}}%
{\color{black}\colorbox{tc22}{\strut \small <|end\_header\_id|>}}%
\end{tabular}
}
\end{minipage}
\hfill
\begin{minipage}[t]{0.3\textwidth}
\centering
\footnotesize
\textbf{Capital prompt}\par
\vspace{0.8em}

\textbf{Cosine similarity}\par\vspace{1.1em}
\resizebox{\linewidth}{!}{%
\input{figures/weighted_texts/token_weights.capital.similarities}%
}

\vspace{2.0em}
\textbf{W-MLP weights}\par\vspace{1.1em}
\resizebox{\linewidth}{!}{%
\input{figures/weighted_texts/token_weights.capital.weights}%
}

\vspace{2.35em}
\textbf{KL divergence}\par\vspace{1.5em}
\resizebox{\linewidth}{!}{%
\input{figures/weighted_texts/token_weights.capital.kl_divergences}%
}
\end{minipage}
\caption{Comparison between the capital prompt with and without chat-template tokens. The short prompt is aligned with the natural-language task sentence inside the full prompt.}
\label{fig:token:capital-full-vs-short}
\end{figure*}

\begin{figure*}[htbp]
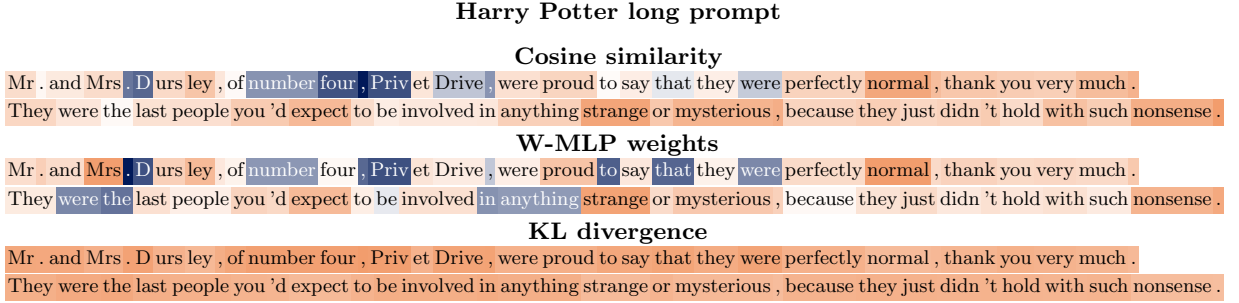

\centering
\begin{minipage}{1\textwidth}
\centering
\footnotesize
\textbf{Harry Potter long prompt}\\
\vspace{0.5em}

\vspace{0.25em}
\textbf{Cosine similarity}
\hspace*{-5.5cm}%
\resizebox{1.35\linewidth}{!}{%
\setlength{\fboxsep}{1pt}

\definecolor{tc0}{RGB}{250,221,204}
\definecolor{tc1}{RGB}{254,247,243}
\definecolor{tc2}{RGB}{252,234,223}
\definecolor{tc3}{RGB}{250,220,203}
\definecolor{tc4}{RGB}{104,118,156}
\definecolor{tc5}{RGB}{86,102,144}
\definecolor{tc6}{RGB}{250,218,200}
\definecolor{tc7}{RGB}{247,197,168}
\definecolor{tc8}{RGB}{252,230,218}
\definecolor{tc9}{RGB}{254,249,247}
\definecolor{tc10}{RGB}{135,147,177}
\definecolor{tc11}{RGB}{64,82,130}
\definecolor{tc12}{RGB}{1,25,89}
\definecolor{tc13}{RGB}{61,80,128}
\definecolor{tc14}{RGB}{252,230,217}
\definecolor{tc15}{RGB}{191,197,213}
\definecolor{tc16}{RGB}{137,148,178}
\definecolor{tc17}{RGB}{251,224,209}
\definecolor{tc18}{RGB}{249,213,192}
\definecolor{tc19}{RGB}{254,251,249}
\definecolor{tc20}{RGB}{253,237,229}
\definecolor{tc21}{RGB}{229,232,238}
\definecolor{tc22}{RGB}{253,238,230}
\definecolor{tc23}{RGB}{197,203,217}
\definecolor{tc24}{RGB}{249,211,189}
\definecolor{tc25}{RGB}{243,167,124}
\definecolor{tc26}{RGB}{246,189,157}
\definecolor{tc27}{RGB}{248,199,171}
\definecolor{tc28}{RGB}{248,204,179}
\definecolor{tc29}{RGB}{249,207,183}
\definecolor{tc30}{RGB}{246,189,156}
\definecolor{tc31}{RGB}{245,178,141}
\definecolor{tc32}{RGB}{250,215,195}
\definecolor{tc33}{RGB}{252,230,218}
\definecolor{tc34}{RGB}{254,249,246}
\definecolor{tc35}{RGB}{251,226,211}
\definecolor{tc36}{RGB}{251,223,208}
\definecolor{tc37}{RGB}{248,203,178}
\definecolor{tc38}{RGB}{248,203,177}
\definecolor{tc39}{RGB}{246,185,151}
\definecolor{tc40}{RGB}{250,221,204}
\definecolor{tc41}{RGB}{253,237,228}
\definecolor{tc42}{RGB}{252,232,221}
\definecolor{tc43}{RGB}{252,231,219}
\definecolor{tc44}{RGB}{249,208,185}
\definecolor{tc45}{RGB}{243,165,121}
\definecolor{tc46}{RGB}{248,200,173}
\definecolor{tc47}{RGB}{245,180,143}
\definecolor{tc48}{RGB}{244,174,135}
\definecolor{tc49}{RGB}{249,212,191}
\definecolor{tc50}{RGB}{249,206,182}
\definecolor{tc51}{RGB}{250,215,195}
\definecolor{tc52}{RGB}{249,210,188}
\definecolor{tc53}{RGB}{248,204,178}
\definecolor{tc54}{RGB}{250,219,201}
\definecolor{tc55}{RGB}{247,193,162}
\definecolor{tc56}{RGB}{248,205,180}
\definecolor{tc57}{RGB}{244,174,134}
\definecolor{tc58}{RGB}{242,157,109}

\noindent
\begin{tabular}{@{}l@{}}
{\color{black}\colorbox{tc0}{\strut \small Mr}}%
{\color{black}\colorbox{tc1}{\strut \small .}}%
{\color{black}\colorbox{tc2}{\strut \small  and}}%
{\color{black}\colorbox{tc3}{\strut \small  Mrs}}%
{\color{white}\colorbox{tc4}{\strut \small .}}%
{\color{white}\colorbox{tc5}{\strut \small  D}}%
{\color{black}\colorbox{tc6}{\strut \small urs}}%
{\color{black}\colorbox{tc7}{\strut \small ley}}%
{\color{black}\colorbox{tc8}{\strut \small ,}}%
{\color{black}\colorbox{tc9}{\strut \small  of}}%
{\color{white}\colorbox{tc10}{\strut \small  number}}%
{\color{white}\colorbox{tc11}{\strut \small  four}}%
{\color{white}\colorbox{tc12}{\strut \small ,}}%
{\color{white}\colorbox{tc13}{\strut \small  Priv}}%
{\color{black}\colorbox{tc14}{\strut \small et}}%
{\color{black}\colorbox{tc15}{\strut \small  Drive}}%
{\color{white}\colorbox{tc16}{\strut \small ,}}%
{\color{black}\colorbox{tc17}{\strut \small  were}}%
{\color{black}\colorbox{tc18}{\strut \small  proud}}%
{\color{black}\colorbox{tc19}{\strut \small  to}}%
{\color{black}\colorbox{tc20}{\strut \small  say}}%
{\color{black}\colorbox{tc21}{\strut \small  that}}%
{\color{black}\colorbox{tc22}{\strut \small  they}}%
{\color{black}\colorbox{tc23}{\strut \small  were}}%
{\color{black}\colorbox{tc24}{\strut \small  perfectly}}%
{\color{black}\colorbox{tc25}{\strut \small  normal}}%
{\color{black}\colorbox{tc26}{\strut \small ,}}%
{\color{black}\colorbox{tc27}{\strut \small  thank}}%
{\color{black}\colorbox{tc28}{\strut \small  you}}%
{\color{black}\colorbox{tc29}{\strut \small  very}}%
{\color{black}\colorbox{tc30}{\strut \small  much}}%
{\color{black}\colorbox{tc31}{\strut \small .}}%
\\[0.2em]
{\color{black}\colorbox{tc32}{\strut \small  They}}%
{\color{black}\colorbox{tc33}{\strut \small  were}}%
{\color{black}\colorbox{tc34}{\strut \small  the}}%
{\color{black}\colorbox{tc35}{\strut \small  last}}%
{\color{black}\colorbox{tc36}{\strut \small  people}}%
{\color{black}\colorbox{tc37}{\strut \small  you}}%
{\color{black}\colorbox{tc38}{\strut \small 'd}}%
{\color{black}\colorbox{tc39}{\strut \small  expect}}%
{\color{black}\colorbox{tc40}{\strut \small  to}}%
{\color{black}\colorbox{tc41}{\strut \small  be}}%
{\color{black}\colorbox{tc42}{\strut \small  involved}}%
{\color{black}\colorbox{tc43}{\strut \small  in}}%
{\color{black}\colorbox{tc44}{\strut \small  anything}}%
{\color{black}\colorbox{tc45}{\strut \small  strange}}%
{\color{black}\colorbox{tc46}{\strut \small  or}}%
{\color{black}\colorbox{tc47}{\strut \small  mysterious}}%
{\color{black}\colorbox{tc48}{\strut \small ,}}%
{\color{black}\colorbox{tc49}{\strut \small  because}}%
{\color{black}\colorbox{tc50}{\strut \small  they}}%
{\color{black}\colorbox{tc51}{\strut \small  just}}%
{\color{black}\colorbox{tc52}{\strut \small  didn}}%
{\color{black}\colorbox{tc53}{\strut \small 't}}%
{\color{black}\colorbox{tc54}{\strut \small  hold}}%
{\color{black}\colorbox{tc55}{\strut \small  with}}%
{\color{black}\colorbox{tc56}{\strut \small  such}}%
{\color{black}\colorbox{tc57}{\strut \small  nonsense}}%
{\color{black}\colorbox{tc58}{\strut \small .}}
\end{tabular}
}

\vspace{0.25em}
\textbf{W-MLP weights}
\hspace*{-5.5cm}%
\resizebox{1.35\linewidth}{!}{%
\setlength{\fboxsep}{1pt}

\definecolor{tc0}{RGB}{251,222,206}
\definecolor{tc1}{RGB}{249,207,184}
\definecolor{tc2}{RGB}{250,220,203}
\definecolor{tc3}{RGB}{242,158,111}
\definecolor{tc4}{RGB}{1,25,89}
\definecolor{tc5}{RGB}{63,81,130}
\definecolor{tc6}{RGB}{249,208,185}
\definecolor{tc7}{RGB}{246,186,152}
\definecolor{tc8}{RGB}{252,231,220}
\definecolor{tc9}{RGB}{253,243,236}
\definecolor{tc10}{RGB}{140,151,180}
\definecolor{tc11}{RGB}{254,248,244}
\definecolor{tc12}{RGB}{88,104,146}
\definecolor{tc13}{RGB}{61,80,128}
\definecolor{tc14}{RGB}{251,228,215}
\definecolor{tc15}{RGB}{252,235,226}
\definecolor{tc16}{RGB}{193,199,215}
\definecolor{tc17}{RGB}{253,238,230}
\definecolor{tc18}{RGB}{247,196,167}
\definecolor{tc19}{RGB}{78,95,139}
\definecolor{tc20}{RGB}{252,231,220}
\definecolor{tc21}{RGB}{85,101,144}
\definecolor{tc22}{RGB}{253,242,235}
\definecolor{tc23}{RGB}{124,136,169}
\definecolor{tc24}{RGB}{250,219,201}
\definecolor{tc25}{RGB}{242,157,109}
\definecolor{tc26}{RGB}{250,218,200}
\definecolor{tc27}{RGB}{249,210,189}
\definecolor{tc28}{RGB}{250,218,199}
\definecolor{tc29}{RGB}{250,215,196}
\definecolor{tc30}{RGB}{250,218,200}
\definecolor{tc31}{RGB}{248,201,174}
\definecolor{tc32}{RGB}{252,235,224}
\definecolor{tc33}{RGB}{122,135,168}
\definecolor{tc34}{RGB}{100,114,154}
\definecolor{tc35}{RGB}{251,224,209}
\definecolor{tc36}{RGB}{253,238,230}
\definecolor{tc37}{RGB}{249,211,190}
\definecolor{tc38}{RGB}{250,217,198}
\definecolor{tc39}{RGB}{246,186,152}
\definecolor{tc40}{RGB}{253,238,229}
\definecolor{tc41}{RGB}{231,233,239}
\definecolor{tc42}{RGB}{251,224,209}
\definecolor{tc43}{RGB}{134,146,176}
\definecolor{tc44}{RGB}{143,154,182}
\definecolor{tc45}{RGB}{243,163,117}
\definecolor{tc46}{RGB}{250,219,202}
\definecolor{tc47}{RGB}{248,200,174}
\definecolor{tc48}{RGB}{249,209,187}
\definecolor{tc49}{RGB}{254,248,245}
\definecolor{tc50}{RGB}{251,227,213}
\definecolor{tc51}{RGB}{250,219,201}
\definecolor{tc52}{RGB}{252,232,221}
\definecolor{tc53}{RGB}{251,225,210}
\definecolor{tc54}{RGB}{252,229,217}
\definecolor{tc55}{RGB}{251,225,210}
\definecolor{tc56}{RGB}{252,234,224}
\definecolor{tc57}{RGB}{245,180,143}
\definecolor{tc58}{RGB}{244,169,127}

\noindent
\begin{tabular}{@{}l@{}}
{\color{black}\colorbox{tc0}{\strut \small Mr}}%
{\color{black}\colorbox{tc1}{\strut \small .}}%
{\color{black}\colorbox{tc2}{\strut \small  and}}%
{\color{black}\colorbox{tc3}{\strut \small  Mrs}}%
{\color{white}\colorbox{tc4}{\strut \small .}}%
{\color{white}\colorbox{tc5}{\strut \small  D}}%
{\color{black}\colorbox{tc6}{\strut \small urs}}%
{\color{black}\colorbox{tc7}{\strut \small ley}}%
{\color{black}\colorbox{tc8}{\strut \small ,}}%
{\color{black}\colorbox{tc9}{\strut \small  of}}%
{\color{white}\colorbox{tc10}{\strut \small  number}}%
{\color{black}\colorbox{tc11}{\strut \small  four}}%
{\color{white}\colorbox{tc12}{\strut \small ,}}%
{\color{white}\colorbox{tc13}{\strut \small  Priv}}%
{\color{black}\colorbox{tc14}{\strut \small et}}%
{\color{black}\colorbox{tc15}{\strut \small  Drive}}%
{\color{black}\colorbox{tc16}{\strut \small ,}}%
{\color{black}\colorbox{tc17}{\strut \small  were}}%
{\color{black}\colorbox{tc18}{\strut \small  proud}}%
{\color{white}\colorbox{tc19}{\strut \small  to}}%
{\color{black}\colorbox{tc20}{\strut \small  say}}%
{\color{white}\colorbox{tc21}{\strut \small  that}}%
{\color{black}\colorbox{tc22}{\strut \small  they}}%
{\color{white}\colorbox{tc23}{\strut \small  were}}%
{\color{black}\colorbox{tc24}{\strut \small  perfectly}}%
{\color{black}\colorbox{tc25}{\strut \small  normal}}%
{\color{black}\colorbox{tc26}{\strut \small ,}}%
{\color{black}\colorbox{tc27}{\strut \small  thank}}%
{\color{black}\colorbox{tc28}{\strut \small  you}}%
{\color{black}\colorbox{tc29}{\strut \small  very}}%
{\color{black}\colorbox{tc30}{\strut \small  much}}%
{\color{black}\colorbox{tc31}{\strut \small .}}%
\\[0.2em]
{\color{black}\colorbox{tc32}{\strut \small  They}}%
{\color{white}\colorbox{tc33}{\strut \small  were}}%
{\color{white}\colorbox{tc34}{\strut \small  the}}%
{\color{black}\colorbox{tc35}{\strut \small  last}}%
{\color{black}\colorbox{tc36}{\strut \small  people}}%
{\color{black}\colorbox{tc37}{\strut \small  you}}%
{\color{black}\colorbox{tc38}{\strut \small 'd}}%
{\color{black}\colorbox{tc39}{\strut \small  expect}}%
{\color{black}\colorbox{tc40}{\strut \small  to}}%
{\color{black}\colorbox{tc41}{\strut \small  be}}%
{\color{black}\colorbox{tc42}{\strut \small  involved}}%
{\color{white}\colorbox{tc43}{\strut \small  in}}%
{\color{white}\colorbox{tc44}{\strut \small  anything}}%
{\color{black}\colorbox{tc45}{\strut \small  strange}}%
{\color{black}\colorbox{tc46}{\strut \small  or}}%
{\color{black}\colorbox{tc47}{\strut \small  mysterious}}%
{\color{black}\colorbox{tc48}{\strut \small ,}}%
{\color{black}\colorbox{tc49}{\strut \small  because}}%
{\color{black}\colorbox{tc50}{\strut \small  they}}%
{\color{black}\colorbox{tc51}{\strut \small  just}}%
{\color{black}\colorbox{tc52}{\strut \small  didn}}%
{\color{black}\colorbox{tc53}{\strut \small 't}}%
{\color{black}\colorbox{tc54}{\strut \small  hold}}%
{\color{black}\colorbox{tc55}{\strut \small  with}}%
{\color{black}\colorbox{tc56}{\strut \small  such}}%
{\color{black}\colorbox{tc57}{\strut \small  nonsense}}%
{\color{black}\colorbox{tc58}{\strut \small .}}%
\end{tabular}
}

\vspace{0.25em}
\textbf{KL divergence}
\hspace*{-5.5cm}%
\resizebox{1.35\linewidth}{!}{%
\setlength{\fboxsep}{1pt}

\definecolor{tc0}{RGB}{243,168,125}
\definecolor{tc1}{RGB}{244,168,126}
\definecolor{tc2}{RGB}{244,168,126}
\definecolor{tc3}{RGB}{245,176,138}
\definecolor{tc4}{RGB}{243,162,117}
\definecolor{tc5}{RGB}{243,162,117}
\definecolor{tc6}{RGB}{245,178,141}
\definecolor{tc7}{RGB}{246,185,150}
\definecolor{tc8}{RGB}{244,174,134}
\definecolor{tc9}{RGB}{243,166,122}
\definecolor{tc10}{RGB}{242,160,114}
\definecolor{tc11}{RGB}{242,160,114}
\definecolor{tc12}{RGB}{242,157,109}
\definecolor{tc13}{RGB}{242,159,112}
\definecolor{tc14}{RGB}{245,178,141}
\definecolor{tc15}{RGB}{243,164,119}
\definecolor{tc16}{RGB}{243,163,117}
\definecolor{tc17}{RGB}{245,177,139}
\definecolor{tc18}{RGB}{244,175,136}
\definecolor{tc19}{RGB}{243,167,124}
\definecolor{tc20}{RGB}{244,169,126}
\definecolor{tc21}{RGB}{243,165,120}
\definecolor{tc22}{RGB}{244,172,131}
\definecolor{tc23}{RGB}{243,164,119}
\definecolor{tc24}{RGB}{245,177,139}
\definecolor{tc25}{RGB}{246,187,154}
\definecolor{tc26}{RGB}{246,184,150}
\definecolor{tc27}{RGB}{245,181,145}
\definecolor{tc28}{RGB}{245,177,138}
\definecolor{tc29}{RGB}{245,179,142}
\definecolor{tc30}{RGB}{245,182,146}
\definecolor{tc31}{RGB}{246,185,151}
\definecolor{tc32}{RGB}{244,176,137}
\definecolor{tc33}{RGB}{244,170,129}
\definecolor{tc34}{RGB}{243,168,126}
\definecolor{tc35}{RGB}{244,175,136}
\definecolor{tc36}{RGB}{245,177,139}
\definecolor{tc37}{RGB}{245,180,143}
\definecolor{tc38}{RGB}{245,181,145}
\definecolor{tc39}{RGB}{246,186,152}
\definecolor{tc40}{RGB}{245,179,142}
\definecolor{tc41}{RGB}{244,174,134}
\definecolor{tc42}{RGB}{244,175,135}
\definecolor{tc43}{RGB}{244,172,131}
\definecolor{tc44}{RGB}{244,175,136}
\definecolor{tc45}{RGB}{246,187,154}
\definecolor{tc46}{RGB}{245,180,144}
\definecolor{tc47}{RGB}{246,185,151}
\definecolor{tc48}{RGB}{246,188,156}
\definecolor{tc49}{RGB}{245,179,142}
\definecolor{tc50}{RGB}{245,177,139}
\definecolor{tc51}{RGB}{245,178,140}
\definecolor{tc52}{RGB}{245,179,142}
\definecolor{tc53}{RGB}{245,179,141}
\definecolor{tc54}{RGB}{245,180,144}
\definecolor{tc55}{RGB}{245,181,145}
\definecolor{tc56}{RGB}{245,180,143}
\definecolor{tc57}{RGB}{246,189,157}
\definecolor{tc58}{RGB}{246,190,159}

\noindent
\begin{tabular}{@{}l@{}}
{\color{black}\colorbox{tc0}{\strut \small Mr}}%
{\color{black}\colorbox{tc1}{\strut \small .}}%
{\color{black}\colorbox{tc2}{\strut \small  and}}%
{\color{black}\colorbox{tc3}{\strut \small  Mrs}}%
{\color{black}\colorbox{tc4}{\strut \small .}}%
{\color{black}\colorbox{tc5}{\strut \small  D}}%
{\color{black}\colorbox{tc6}{\strut \small urs}}%
{\color{black}\colorbox{tc7}{\strut \small ley}}%
{\color{black}\colorbox{tc8}{\strut \small ,}}%
{\color{black}\colorbox{tc9}{\strut \small  of}}%
{\color{black}\colorbox{tc10}{\strut \small  number}}%
{\color{black}\colorbox{tc11}{\strut \small  four}}%
{\color{black}\colorbox{tc12}{\strut \small ,}}%
{\color{black}\colorbox{tc13}{\strut \small  Priv}}%
{\color{black}\colorbox{tc14}{\strut \small et}}%
{\color{black}\colorbox{tc15}{\strut \small  Drive}}%
{\color{black}\colorbox{tc16}{\strut \small ,}}%
{\color{black}\colorbox{tc17}{\strut \small  were}}%
{\color{black}\colorbox{tc18}{\strut \small  proud}}%
{\color{black}\colorbox{tc19}{\strut \small  to}}%
{\color{black}\colorbox{tc20}{\strut \small  say}}%
{\color{black}\colorbox{tc21}{\strut \small  that}}%
{\color{black}\colorbox{tc22}{\strut \small  they}}%
{\color{black}\colorbox{tc23}{\strut \small  were}}%
{\color{black}\colorbox{tc24}{\strut \small  perfectly}}%
{\color{black}\colorbox{tc25}{\strut \small  normal}}%
{\color{black}\colorbox{tc26}{\strut \small ,}}%
{\color{black}\colorbox{tc27}{\strut \small  thank}}%
{\color{black}\colorbox{tc28}{\strut \small  you}}%
{\color{black}\colorbox{tc29}{\strut \small  very}}%
{\color{black}\colorbox{tc30}{\strut \small  much}}%
{\color{black}\colorbox{tc31}{\strut \small .}}%
\\[0.2em]
{\color{black}\colorbox{tc32}{\strut \small  They}}%
{\color{black}\colorbox{tc33}{\strut \small  were}}%
{\color{black}\colorbox{tc34}{\strut \small  the}}%
{\color{black}\colorbox{tc35}{\strut \small  last}}%
{\color{black}\colorbox{tc36}{\strut \small  people}}%
{\color{black}\colorbox{tc37}{\strut \small  you}}%
{\color{black}\colorbox{tc38}{\strut \small 'd}}%
{\color{black}\colorbox{tc39}{\strut \small  expect}}%
{\color{black}\colorbox{tc40}{\strut \small  to}}%
{\color{black}\colorbox{tc41}{\strut \small  be}}%
{\color{black}\colorbox{tc42}{\strut \small  involved}}%
{\color{black}\colorbox{tc43}{\strut \small  in}}%
{\color{black}\colorbox{tc44}{\strut \small  anything}}%
{\color{black}\colorbox{tc45}{\strut \small  strange}}%
{\color{black}\colorbox{tc46}{\strut \small  or}}%
{\color{black}\colorbox{tc47}{\strut \small  mysterious}}%
{\color{black}\colorbox{tc48}{\strut \small ,}}%
{\color{black}\colorbox{tc49}{\strut \small  because}}%
{\color{black}\colorbox{tc50}{\strut \small  they}}%
{\color{black}\colorbox{tc51}{\strut \small  just}}%
{\color{black}\colorbox{tc52}{\strut \small  didn}}%
{\color{black}\colorbox{tc53}{\strut \small 't}}%
{\color{black}\colorbox{tc54}{\strut \small  hold}}%
{\color{black}\colorbox{tc55}{\strut \small  with}}%
{\color{black}\colorbox{tc56}{\strut \small  such}}%
{\color{black}\colorbox{tc57}{\strut \small  nonsense}}%
{\color{black}\colorbox{tc58}{\strut \small .}}%
\end{tabular}
}

\end{minipage}
\caption{Qualitative token-level visualization for a longer prompt, split into two lines for readability.}
\label{fig:harry-potter-token-weights}
\end{figure*}

\paragraph{Qualitative Observations.}
Representative token-level visualizations are shown in Figures~\ref{fig:token-weights-examples-2}, \ref{fig:token-weights-examples-3}, \ref{fig:token:capital-full-vs-short}, and~\ref{fig:harry-potter-token-weights}.  For each prompt, we visualize the cosine similarity to the final patch vector, the KL-divergence-based comparison, and the weights assigned by the W-MLP.

\newcommand{\PromptPanel}[3][\columnwidth]{%
\begin{minipage}[t]{#1}
\centering
\scriptsize
\vspace{0.4em}
\textbf{#3}\par
\vspace{0.15em}

\textbf{Cosine similarity}\par
\input{figures/weighted_texts/token_weights.#2.similarities.tex}\par

\vspace{0.15em}
\textbf{W-MLP weights}\par
\input{figures/weighted_texts/token_weights.#2.weights.tex}\par

\vspace{0.15em}
\textbf{KL divergence}\par
\input{figures/weighted_texts/token_weights.#2.kl_divergences.tex}\par

\end{minipage}%
}

\begin{figure}[h!]
\centering
\begin{minipage}[t]{\columnwidth}
\centering
\scriptsize
\vspace{0.4em}
\textbf{Ocean animals}\par
\vspace{0.15em}

\textbf{Cosine similarity}\par
\input{figures/weighted_texts/token_weights.ocean_animals_prompt.similarities.tex}\par

\vspace{0.15em}
\textbf{W-MLP weights}\par
\input{figures/weighted_texts/token_weights.ocean_animals_prompt.weights.tex}\par

\vspace{0.15em}
\textbf{KL divergence}\par
\input{figures/weighted_texts/token_weights.ocean_animals_prompt.kl_divergences.tex}\par

\end{minipage}%

\vspace{0.5em}
\begin{minipage}[t]{\columnwidth}
\centering
\scriptsize
\vspace{0.4em}
\textbf{Moon question}\par
\vspace{0.15em}

\textbf{Cosine similarity}\par
\input{figures/weighted_texts/token_weights.moon_question.similarities.tex}\par

\vspace{0.15em}
\textbf{W-MLP weights}\par
\input{figures/weighted_texts/token_weights.moon_question.weights.tex}\par

\vspace{0.15em}
\textbf{KL divergence}\par
\input{figures/weighted_texts/token_weights.moon_question.kl_divergences.tex}\par

\end{minipage}%

\vspace{0.5em}
\begin{minipage}[t]{\columnwidth}
\centering
\scriptsize
\vspace{0.4em}
\textbf{German $\to$ Chinese translation}\par
\vspace{0.15em}

\textbf{Cosine similarity}\par
\input{figures/weighted_texts/token_weights.german2chinese_translation.similarities.tex}\par

\vspace{0.15em}
\textbf{W-MLP weights}\par
\input{figures/weighted_texts/token_weights.german2chinese_translation.weights.tex}\par

\vspace{0.15em}
\textbf{KL divergence}\par
\input{figures/weighted_texts/token_weights.german2chinese_translation.kl_divergences.tex}\par

\end{minipage}%

\vspace{-0.9em}
\caption{Qualitative token-level visualizations.}
\label{fig:token-weights-examples-3}
\end{figure}

\newpage

\vspace{-5em}
\paragraph{Interpretation}
Across the examples, the learned weights tend to assign larger values to tokens that are semantically relevant for the task, such as task words, relation words, or answer-format cues. For example, high weights often appear on tokens such as \texttt{capital}, \texttt{authored}, \texttt{animals}, or \texttt{Chinese translation}. This agrees with the intuition from the hand-made weighting experiment: task information is not distributed uniformly across the prompt, but is concentrated in a small number of informative token positions.

The cosine similarity and KL-divergence visualizations provide a complementary view. Tokens near the beginning of the prompt are generally less aligned with the final patch vector, while later and semantically more informative tokens are often closer. The capital full-prompt example also illustrates the effect of the prompt format: in the chat-formatted setting, the task sentence \texttt{What is the capital of this country?} appears together with special template tokens, which can also receive high similarity or weight. This suggests that the learned patch vector can capture both task content and parts of the surrounding prompt structure.

These figures are intended as qualitative evidence, rather than a causal analysis of individual tokens. They provide a useful sanity check that the learned aggregation mechanism tends to focus on tokens that are meaningful for the task, rather than behaving like a uniform average over the prompt.

\vspace{5em}

\section{Training parameters}
\label{app:hyperparameters}
The training and evaluation parameters are summarized in ~\autoref{tab:training-parameters}.

\begin{table}[h!]
\centering
\scriptsize
\begin{tabular}{ll}
\toprule
Parameter & Value \\
\midrule
Base model & Llama-3.1-8B-Instruct \\
Prompt format & instruction/chat template \\
Training tasks & 8 toy task families \\
Extraction layer & $12$ \\
Patching layer & $2$ \\
Patch scaler & $1.0$ \\
Learning rate & $10^{-3}$ \\
Epochs & $5$ \\
Batch size & W-MLP: $8$; Transformer: $16$ \\
Target token position & $6$ \\
Generation length & $20$ \\
Temperature & $0.75$ \\
Placeholder token & \texttt{U+FFFD}, rendered as \texttt{¿} \\
W-MLP arch. & $4096\!\to\!2048\!\to\!1024\!\to\!512\!\to\!256\!\to\!1$ \\
Transformer arch. & $d_{\mathrm{model}}=512$, $2$ heads, $2$ layers \\
Mean centering & W-MLP only \\
\bottomrule
\end{tabular}
\caption{Training and evaluation parameters.}
\label{tab:training-parameters}
\end{table}

\end{document}